\documentclass[twoside,twocolumn,9pt]{article}
\usepackage{bm}
\usepackage{extsizes}
\usepackage[super,sort&compress,comma]{natbib} 
\usepackage[version=3]{mhchem}
\usepackage[left=1.5cm, right=1.5cm, top=1.785cm, bottom=2.0cm]{geometry}
\usepackage{balance}
\usepackage{mathptmx}
\usepackage{sectsty}
\usepackage{graphicx} 
\usepackage{lastpage}
\usepackage[format=plain,justification=justified,singlelinecheck=false,font={stretch=1.125,small,sf},labelfont=bf,labelsep=space]{caption}
\usepackage{float}
\usepackage{fancyhdr}
\usepackage{fnpos}
\usepackage[english]{babel}
\addto{\captionsenglish}{%
  
}
\usepackage{array}
\usepackage{droidsans}
\usepackage{charter}
\usepackage[T1]{fontenc}
\usepackage[usenames,dvipsnames]{xcolor}
\usepackage{setspace}
\usepackage[compact]{titlesec}
\usepackage{hyperref}

\usepackage{epstopdf}
\usepackage[utf8]{inputenc}
\usepackage{textcomp}

\usepackage{enumitem}

\usepackage{booktabs}

\usepackage{multirow}

\usepackage{enumitem}

\usepackage{tikz}
\usepackage{pgfplots}
\pgfplotsset{compat=1.18}   

\usepackage{subcaption}

\usepackage{amssymb,amsmath}

\usepackage{bm}     

\usepackage{balance}

\usetikzlibrary{positioning}
\usetikzlibrary{positioning, calc, arrows.meta, decorations.pathmorphing, patterns}
\usepackage{tcolorbox}
\tcbuselibrary{skins}



\usepackage[utf8]{inputenc}        
\usepackage{pifont}                
\newcommand{\cmark}{\ding{51}}     
\newcommand{\xmark}{\ding{55}}     
\DeclareUnicodeCharacter{2713}{\cmark}  
\DeclareUnicodeCharacter{2717}{\xmark}  

\usepackage[utf8]{inputenc}

\usepackage{xcolor}

\tcbset{
  colback=blue!5!white,
  colframe=blue!60!black,
  boxrule=1pt,
  arc=5pt,
  top=5pt,
  bottom=5pt,
  left=5pt,
  right=5pt,
  boxsep=5pt
}
\usetikzlibrary{arrows.meta,decorations.pathmorphing,calc,patterns}

\definecolor{cream}{RGB}{222,217,201}

\def\methodName{OPENXRD}

\begin{document}

\pagestyle{fancy}
\thispagestyle{plain}
\fancypagestyle{plain}{
\renewcommand{\headrulewidth}{0pt}
}

\makeFNbottom
\makeatletter
\renewcommand\LARGE{\@setfontsize\LARGE{15pt}{17}}
\renewcommand\Large{\@setfontsize\Large{12pt}{14}}
\renewcommand\large{\@setfontsize\large{10pt}{12}}
\renewcommand\footnotesize{\@setfontsize\footnotesize{7pt}{10}}
\makeatother

\renewcommand{\thefootnote}{\fnsymbol{footnote}}
\renewcommand\footnoterule{\vspace*{1pt}%
\color{cream}\hrule width 3.5in height 0.4pt \color{black}\vspace*{5pt}} 
\setcounter{secnumdepth}{5}

\makeatletter 
\renewcommand\@biblabel[1]{#1}            
\renewcommand\@makefntext[1]%
{\noindent\makebox[0pt][r]{\@thefnmark\,}#1}
\makeatother 
\renewcommand{\figurename}{\small{Fig.}~}
\sectionfont{\sffamily\Large}
\subsectionfont{\normalsize}
\subsubsectionfont{\bf}
\setstretch{1.125}
\setlength{\skip\footins}{0.8cm}
\setlength{\footnotesep}{0.25cm}
\setlength{\jot}{10pt}
\titlespacing*{\section}{0pt}{4pt}{4pt}
\titlespacing*{\subsection}{0pt}{15pt}{1pt}

\fancyfoot{}
\fancyfoot[RO]{\footnotesize{\sffamily{\thepage}}}
\fancyfoot[LE]{\footnotesize{\sffamily{\thepage}}}
\fancyhead{}
\renewcommand{\headrulewidth}{0pt} 
\renewcommand{\footrulewidth}{0pt}
\setlength{\arrayrulewidth}{1pt}
\setlength{\columnsep}{6.5mm}
\setlength\bibsep{1pt}

\makeatletter 
\newlength{\figrulesep} 
\setlength{\figrulesep}{0.5\textfloatsep} 

\newcommand{\topfigrule}{\vspace*{-1pt}%
\noindent{\color{cream}\rule[-\figrulesep]{\columnwidth}{1.5pt}} }

\newcommand{\botfigrule}{\vspace*{-2pt}%
\noindent{\color{cream}\rule[\figrulesep]{\columnwidth}{1.5pt}} }

\newcommand{\dblfigrule}{\vspace*{-1pt}%
\noindent{\color{cream}\rule[-\figrulesep]{\textwidth}{1.5pt}} }

\makeatother

\twocolumn[
  \begin{@twocolumnfalse}
\vspace{1em}
\sffamily
\begin{center}
{\LARGE\textbf{OPENXRD: A Comprehensive Benchmark  Framework for \\ LLM/MLLM XRD Question Answering$^\dag$}}

\vspace{0.5cm}

{\large Ali Vosoughi,$^{\ast}$\textit{$^{a}$} Ayoub Shahnazari,$^{\ast}$\textit{$^{b}$} Yufeng Xi,\textit{$^{c}$} Zeliang Zhang,\textit{$^{a}$} Griffin Hess,\textit{$^{b}$} Chenliang Xu\textit{$^{a}$} and Niaz Abdolrahim\textit{$^{b,c,d}$}}

\vspace{0.5cm}

\begin{minipage}{0.9\textwidth}
\normalsize
We introduce OPENXRD, a comprehensive benchmarking framework for evaluating large language models (LLMs) and multimodal LLMs (MLLMs) in crystallography question answering. The framework measures context assimilation, or how models use fixed, domain-specific supporting information during inference. The framework includes 217 expert-curated X-ray diffraction (XRD) questions covering fundamental to advanced crystallographic concepts, each evaluated under closed-book (without context) and open-book (with context) conditions, where the latter includes concise reference passages generated by GPT-4.5 and refined by crystallography experts. We benchmark 74 state-of-the-art LLMs and MLLMs, including GPT-4, GPT-5, O-series, LLaVA, LLaMA, QWEN, Mistral, and Gemini families, to quantify how different architectures and scales assimilate external knowledge. Results show that mid-sized models (7B–70B parameters) gain the most from contextual materials, while very large models often show saturation or interference and the largest relative gains appear in small and mid-sized models. Expert-reviewed materials provide significantly higher improvements than AI-generated ones even when token counts are matched, confirming that content quality, not quantity, drives performance. OPENXRD offers a reproducible diagnostic benchmark for assessing reasoning, knowledge integration, and guidance sensitivity in scientific domains, and provides a foundation for future multimodal and retrieval-augmented crystallography systems.
\end{minipage}
\end{center}

\end{@twocolumnfalse} \vspace{0.6cm}
]

\renewcommand*\rmdefault{bch}\normalfont\upshape
\rmfamily
\section*{}
\vspace{-1cm}

\footnotetext{$^{\ast}$~These authors contributed equally to this work.}

\footnotetext{\textit{$^{a}$~Dept. of Computer Science, University of Rochester, Rochester, New York 14627, USA.}}
\footnotetext{\textit{$^{b}$~Dept. of Mechanical Engineering, University of Rochester, Rochester, 14627, NY, USA. }}
\footnotetext{\textit{$^{c}$~Material Science Program, University of Rochester, Rochester, 14627, NY, USA. }}
\footnotetext{\textit{$^{d}$~Staff Scientist, Laboratory for Laser Energetics (LLE), University of Rochester, Rochester, 14627, NY, USA. }}

\footnotetext{\dag~ The codes and benchmarking dataset can be accessed at\\ \url{https://github.com/niaz60/OpenXRD}. \\Project webpage: \url{https://niaz60.github.io/OpenXRD/}.}

\section{Introduction}\label{sec:intro}

Crystallography is the scientific discipline concerned with determining the arrangement of atoms and molecules in crystalline solids \cite{he2018two, rathore2021advancements, ubic2024crystallography}. This structural understanding is crucial for elucidating material properties, such as symmetry, geometry, and physical characteristics \cite{vainshtein2013fundamentals, muller2024symmetry}. This understanding is crucial for progressing materials science, especially in areas like metallurgy, pharmaceuticals, and semiconductor technology \cite{raabe2023materials, altuntacs2023metallurgical, bijak2023current, afraj2022heteroalkyl, inoue2021emerging}. Central to crystallography is X-ray diffraction (XRD), a key experimental technique that enables researchers to uncover detailed information about crystal structures by examining how X-rays interact with crystalline lattices \cite{cullity1957elements, ali2022x, bunaciu2015x}.

The foundational principle of XRD is Bragg's Law, discovered by Lawrence Bragg in 1912 \cite{cantor2020equations, bragg1968x}. Bragg's law describes how the angle of a diffracted X-ray beam depends on the wavelength of the X-rays and the spacing between atoms and molecules within the material \cite{pope1997x, stohr2023nature}. Using XRD, researchers analyze crystal structures to determine important parameters such as atomic arrangements, phase composition, unit cell size, grain size, crystallinity, strain, and lattice defects \cite{fatimah2022calculate, lee2021data, uvarov2019influence, he2018method, qadr2023calculating, dolabella2022lattice}.

Advances in the analysis of large XRD data have demonstrated that conventional deep learning approaches, such as Convolutional Neural Networks (CNNs) and Graph Neural Networks (GNNs), are powerful tools, particularly for identifying or predicting crystal parameters space group labels \cite{salgado2023automated, ziletti2018insightful, park2017classification}, lattice constants \cite{gomez2023convolutional, chakraborty2022deep}, or phase compositions \cite{lee2020deep, szymanski2023adaptively}. These models achieve remarkable numerical accuracy, demonstrating their efficacy for crystal classification and structural analysis tasks.

However, despite their success in quantitative predictions, these deep-learning techniques are limited in providing interpretative and explanatory insights into the underlying physics or chemistry of XRD data \cite{faber2017prediction, surdu2023x, lee2023deep}. This lack of interpretability restricts their ability to offer meaningful explanations about the material properties or structural characteristics encoded within the diffraction data. This lack of interpretability restricts the broader application of deep-learning approaches in crystallographic research, where understanding underlying phenomena is essential \cite{wu2021three, omori2023recent, chavez2025applied}.

In contrast, recent progress in Natural Language Processing (NLP) driven by Large Language Models (LLMs), such as GPT-based architectures, has transformed various computational domains \cite{liu2023summary, hadi2023large, dahl2024large}. LLMs demonstrate exceptional capabilities in open-ended question answering \cite{zhuang2023toolqa, kamalloo2023evaluating}, multi-step reasoning \cite{chen2024optimizing, patel2024multi}, domain-specific question answering \cite{zaki2024mascqa, balhorn2023does, mirza2024large}, predicting material properties \cite{choudhary2025microscopygpt,rubungo2023llm, qian2023can}, and extracting information from complex datasets \cite{li2025hybrid,chiang2024llamp, gupta2022matscibert}. These powerful language models can potentially bridge the interpretability gap faced by traditional deep-learning models when applied to scientific fields, including materials science and crystallography \cite{yu2024large, shi2025comparison, wang2025evaluating}.

For instance, Antunes \emph{et al.}\ introduced CrystaLLM, an autoregressive LLM trained on millions of Crystallographic Information Files (CIFs), which encode detailed crystallographic data, including atomic coordinates, symmetry operations, lattice parameters, and space group information, to generate plausible crystal structures from given chemical compositions, demonstrating the potential of text-based crystal structure generation~\cite{Antunes2024CrystaLLM}. In addition, Johansen et al. introduced deCIFer, an autoregressive language model on powder XRD data to produce complete crystal structures in CIF format, attaining high match rates between generated structures and experimental diffraction profiles~\cite{Johansen2025deCIFer}. 

Further extending these capabilities, Choudhary introduced DiffractGPT, a generative pretrained transformer that predicts atomic structures directly from XRD patterns, particularly when chemical information is provided~\cite{choudhary2025diffractgpt}. AtomGPT, another transformer-based model by Choudhary, effectively predicts material properties such as formation energies, electronic bandgaps, and superconducting transition temperatures with accuracy comparable to GNNs~\cite{choudhary2024atomgpt}. Beyond generative tasks, LLMs have also been employed for scientific question-answering and knowledge retrieval in the materials domain. For example, the LLaMP framework combines an LLM with a materials database to fetch and reason over crystallographic data, reducing hallucinations and improving factual accuracy in materials science QA \cite{chiang2024llamp}.

In this study, we investigate whether providing domain-specific context generated by a stronger model, such as supporting textual material, can significantly enhance the performance of weaker models or alternative configurations of the same model on specialized XRD-related questions. To rigorously test this hypothesis, we constructed a carefully selected dataset consisting of 217 multiple-choice XRD questions, each reviewed and approved by a domain expert at the Ph.D. level. Each question has only one correct answer among four provided choices, covering a range of topics from fundamental principles to complex scenarios, including issues such as basic structural geometry, unit cell dimensions, and coordination environments, to more complex concepts, including fundamental equations and symmetry analysis. Initially, we evaluate the performance of various Large LLMs, including GPT-4.5, GPT-4, O1, O3, and so forth, under closed-book conditions, where models rely solely on their pretrained internal knowledge without external assistance. Our initial results show that GPT-4.5 clearly outperforms other tested models. Based on this outcome, we employ GPT-4.5 to generate approximately one-page textual summaries. These summaries intentionally avoid explicitly stating correct answers yet offer adequate context and guidance to facilitate correct reasoning. We then reassess the performance of the other models under open‑book conditions to quantify the improvement when provided with these contextual summaries, aiming to quantify the improvement enabled by this supplementary textual material.

It is important to clarify that OPENXRD is a context-assimilation benchmark rather than a retrieval system, and as such is complementary to Retrieval-Augmented Generation (RAG) rather than competing with it. While RAG is a deployment architecture that couples a retriever with a generator to fetch answer-bearing passages from large corpora at inference time, OPENXRD deliberately removes the retrieval confound by providing fixed, curated, answer-guiding passages that avoid revealing correct answers. Accordingly, our open-book setting should be interpreted as an oracle (``Gold-Standard'') RAG condition: the generator receives perfectly relevant, expert-verified context (i.e., perfect retrieval), so the measured accuracy is an upper bound on what an end-to-end RAG system could achieve with the same base model under ideal retrieval. This design isolates the language model's ability to integrate external guidance from confounding factors such as retrieval quality, chunking, and ranking. OPENXRD does not perform retrieval and does not aspire to be a deployable QA system; instead, it provides a controlled, reproducible setting to disclose how language models react to guidance—when it helps, when it distracts, and how sensitivity varies with model family, token budget, and content quality. These are properties that end-to-end RAG often obscure because retrieved passages can directly contain the answer. The same evaluation harness can benchmark RAG systems by replacing our oracle helper passages with retrieved chunks from a crystallography corpus, thereby decomposing end-to-end RAG accuracy into retrieval quality and language model assimilation capability. Thus, OPENXRD serves as a diagnostic complement for RAG research, not a competing alternative.

\section{Methods}
\label{sec:method}

\subsection{Dataset Selection and Curation}

\begin{figure}[ht]
   \centering
   \includegraphics[width=\linewidth]{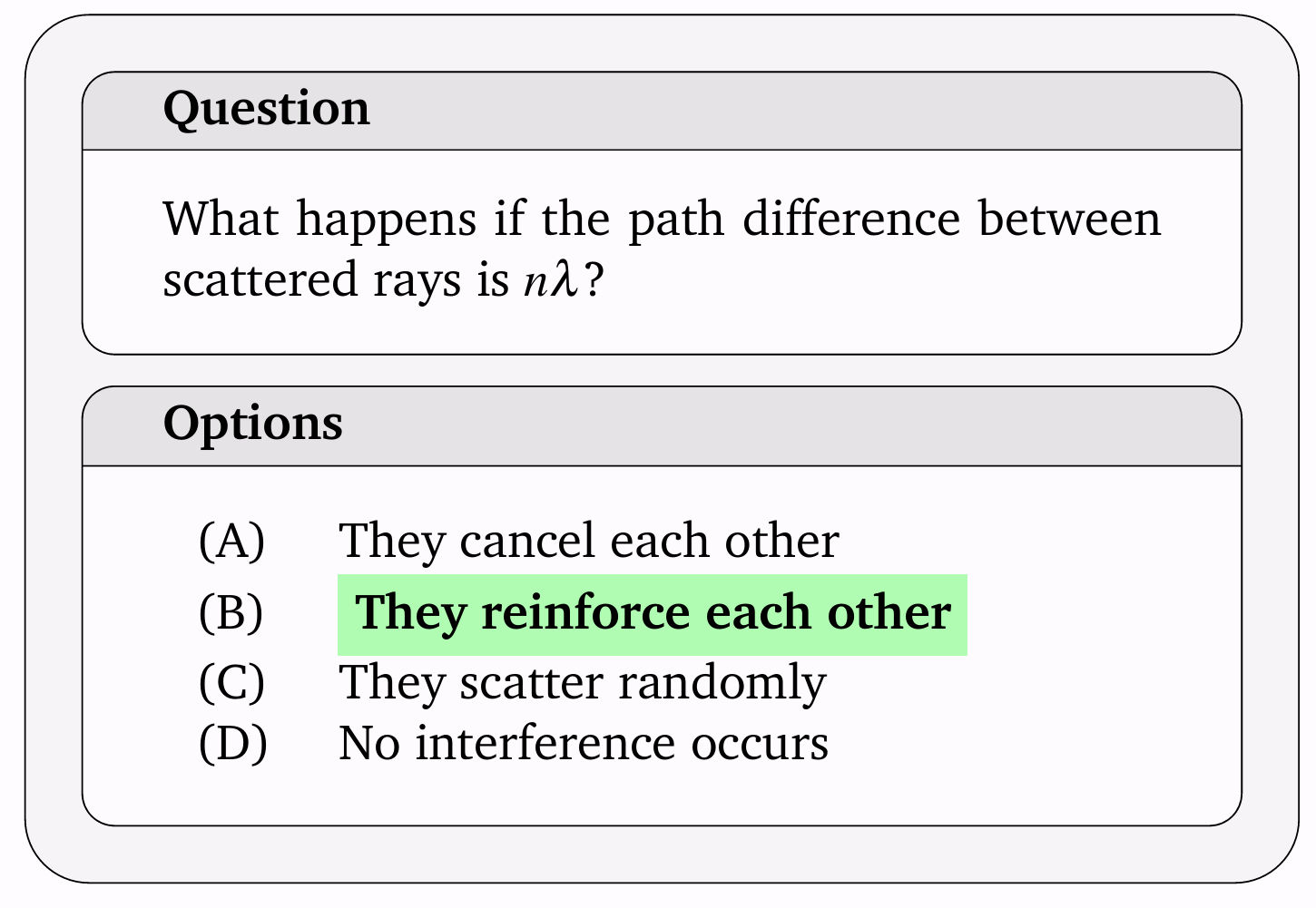}
   \caption{Example question from the OPENXRD dataset in closed-book format. Each question includes a concise prompt, multiple-choice options (3-4 answers), with one correct answer. This illustrates the baseline evaluation condition where models must rely solely on their internal knowledge without external supporting materials.}
   \label{fig:example_question_closed}
\end{figure}

\begin{figure}[ht]
    \centering
    \includegraphics[width=\linewidth]{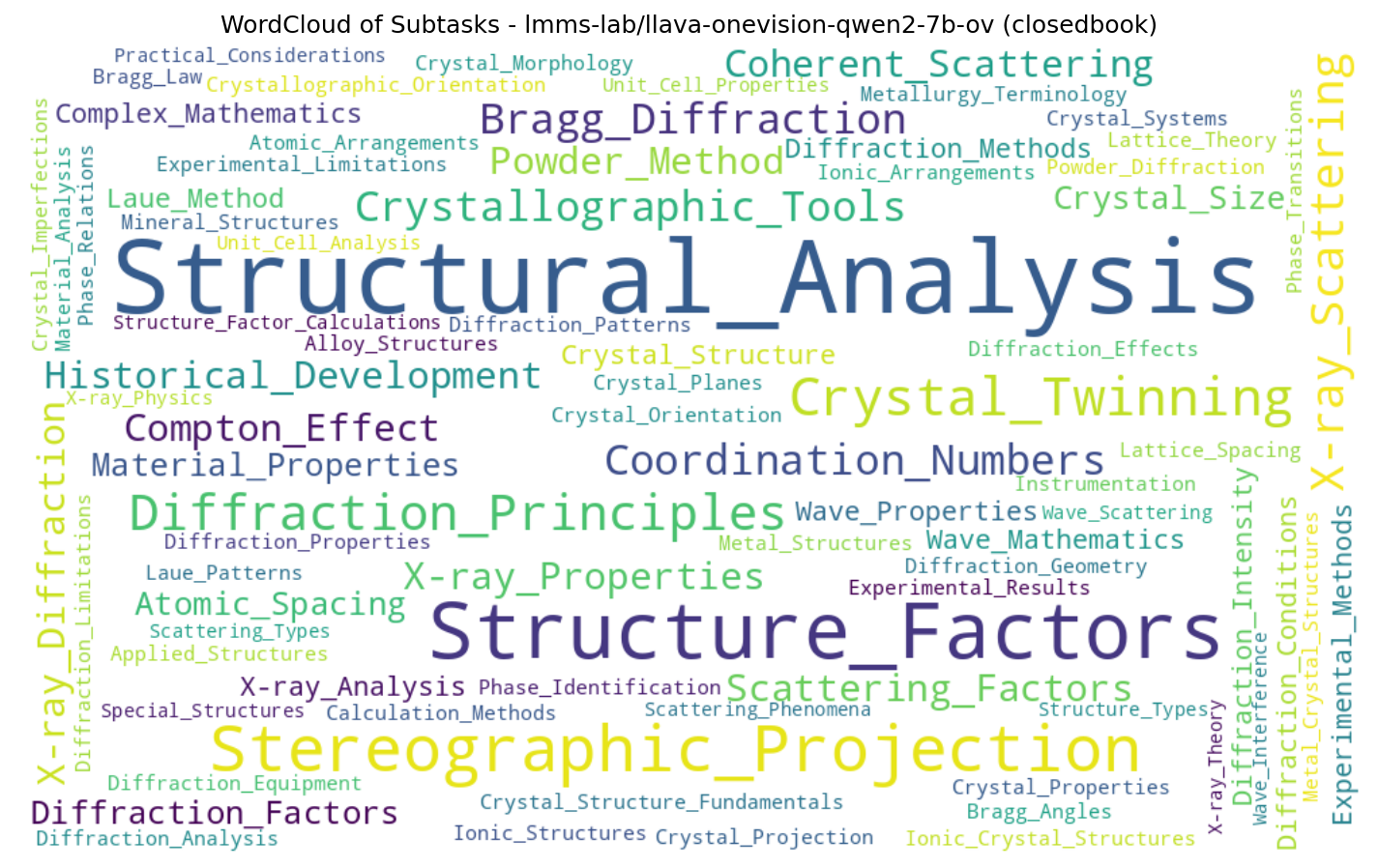}
    \caption{A word cloud of the \emph{subtask} labels from our human-curated crystallography dataset.
    Larger words indicate subtasks with a higher number of questions, illustrating the breadth of topics
    (e.g., diffraction fundamentals, lattice geometry, advanced structure analysis). This distribution
    underscores the diversity and domain complexity captured in our QA benchmark, spanning both basic
    definitions (e.g., counting crystal systems) and intricate reflections (e.g., twin boundaries, space-group
    anomalies).}
    \label{fig:wordcloud_subtasks}
\end{figure}

To evaluate language model performance in crystallography, we curated a domain-specific multiple-choice question-answering dataset comprising 217 expert-reviewed items focused exclusively on X-ray diffraction and crystallographic reasoning. To construct this dataset, we partially drew on content from Cullity's Introduction to X-ray Diffraction \cite{cullity1957elements}. Each question  includes a concise prompt,  three to four answer options, a single correct answer, a brief explanation providing the scientific rationale, and a associated subtask label that enables fine-grained performance analysis across different crystallographic domains.  The dataset encompasses 81 distinct crystallographic subtasks, covering a broad span of domain knowledge including crystal structure fundamentals, diffraction geometry, reciprocal-space concepts, Bragg-law applications, peak indexing, space-group interpretation, unit-cell analysis, scattering physics, microstructure effects, and additional specialized topics (see Figure~\ref{fig:example_question_closed} and Figure~\ref{fig:wordcloud_subtasks}). 
We adopt a multiple-choice format in this foundational benchmark because it enables unambiguous, reproducible scoring across many models, avoiding the evaluator subjectivity that arises in open-ended generation. This design choice is intended to isolate \emph{context assimilation} effects under controlled conditions rather than to fully represent real-world crystallographic workflows. Evaluating open-ended, problem-solving tasks (e.g., short-answer explanations and multi-step derivations) is an important direction but is beyond the scope of the present benchmark and is discussed as future work.

As illustrated in Figure~\ref{fig:wordcloud_subtasks}, the distribution of subtask labels highlights the diversity and complexity of the topics covered in our dataset, ranging from basic definitions to advanced structural phenomena. All questions were reviewed and validated by a domain expert to ensure scientific correctness and comprehensive coverage of essential XRD subtasks.

This annotated dataset serves three primary purposes. First, it provides a benchmarking framework for testing the capability of LLMs to comprehend and apply specialized crystallographic concepts. Second, it enables fine-grained subtask-level analysis, distinguishing model performance across different categories, ranging from fundamental definitions to complex conditions. Third, it supports a comparative study between closed-book and open-book evaluation settings, allowing us to assess the impact of supplementary domain-specific context on model accuracy. 

\begin{figure}[ht]
\centering
\begin{tcolorbox}[
    enhanced,
    colback=gray!8,
    colframe=black,
    arc=4mm,
    width=0.9\linewidth,
    boxrule=0.3pt,
    fontupper=\normalsize\normalfont,
    fontlower=\normalsize\normalfont
]
    \begin{tcolorbox}[
        colback=white,
        colframe=black,
        title={\normalsize\bfseries AI-Generated Material},
        fonttitle=\normalsize\bfseries,
        fontupper=\normalsize\normalfont,
        coltitle=black,
        colbacktitle=gray!25,
        arc=2mm,
        boxrule=0.3pt
    ]
        \normalsize\normalfont The atomic scattering factor measures how effectively an atom scatters X-rays, and it depends on the atom's electron density and the scattering angle.
    \end{tcolorbox}
    \begin{tcolorbox}[
        colback=white,
        colframe=black,
        title={\normalsize\bfseries Expert-Reviewed Material},
        fonttitle=\normalsize\bfseries,
        fontupper=\normalsize\normalfont,
        coltitle=black,
        colbacktitle=gray!25,
        arc=2mm,
        boxrule=0.3pt,
        left=4pt,
        right=4pt,
        top=4pt,
        bottom=4pt
    ]
        \normalsize\normalfont The atomic scattering factor quantifies how effectively an atom scatters X-rays, depending primarily on its electron density distribution and the scattering angle. At low scattering angles, the atomic scattering factor approaches the total electron count of the atom, but as the angle increases, it decreases due to interference effects within the electron cloud. Consequently, heavier atoms with higher electron counts typically have larger scattering factors at low angles. Additionally, this scattering factor influences the intensity of diffraction peaks, directly affecting how clearly atomic positions can be determined from experimental X-ray diffraction data.
    \end{tcolorbox}
\end{tcolorbox}
\caption{The expert-reviewed version provides essential clarifications about the angular dependence of atomic scattering factors, emphasizes their direct relationship to atomic electron count, and highlights practical implications for interpreting diffraction data, elements often missed in AI-generated explanations.}
\label{fig:expert-reviewed}
\end{figure}

\begin{figure}[ht]
   \centering
   \includegraphics[width=\linewidth]{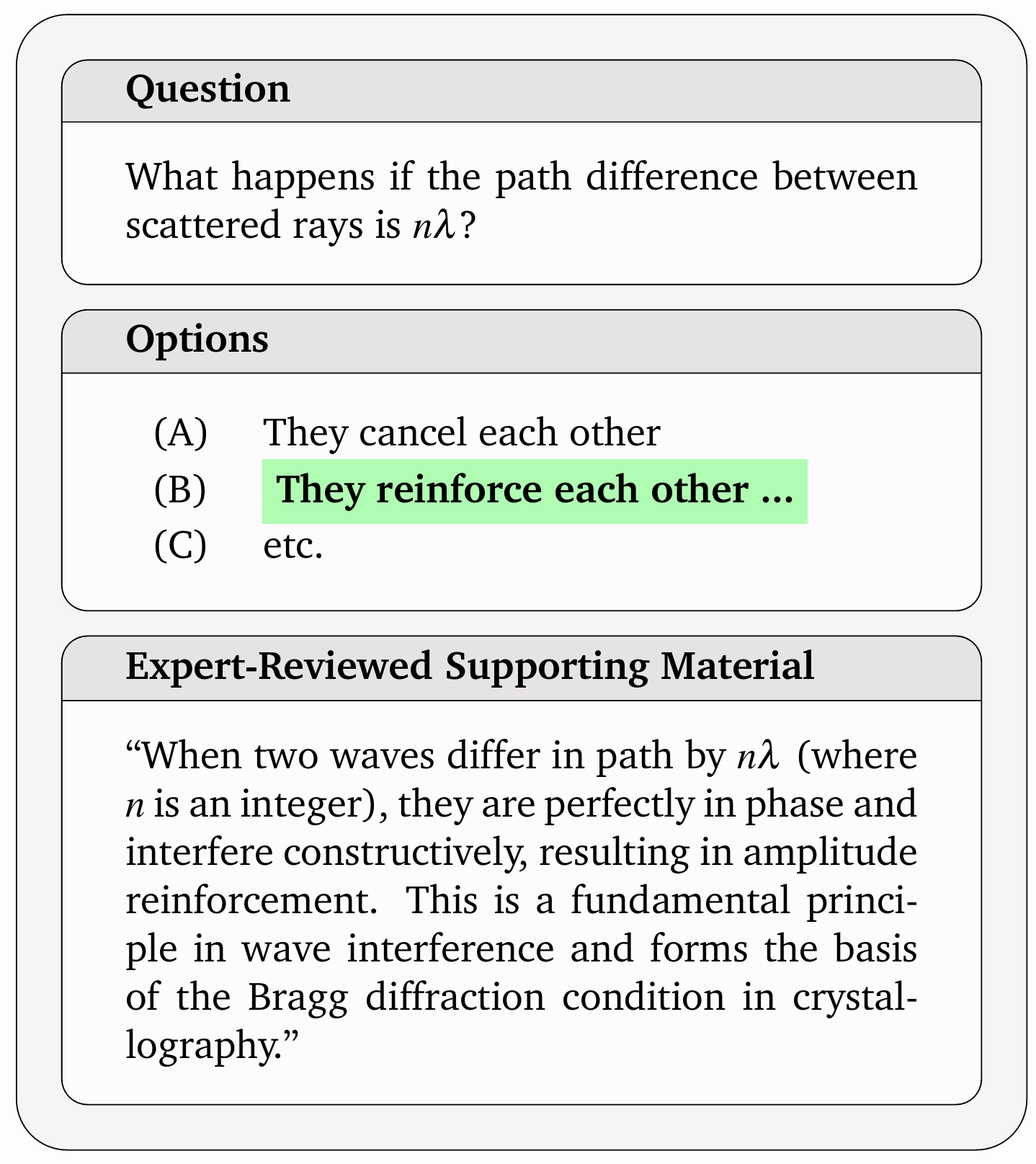}
   \caption{An open-book mode example with expert-reviewed supporting material that clearly explains how a path difference of $n\lambda$ leads to constructive interference, with additional contextual information about its relevance to crystallography. The correct answer is highlighted in green.}
   \label{fig:qa_example}
\end{figure}


\subsection{Evaluation Framework: Closed-Book vs. Open-Book}
The OPENXRD evaluation framework is designed to assess the reasoning capabilities of language models in crystallography under two distinct conditions: closed-book and open-book modes.
In the closed-book mode, the model receives only the question along with its corresponding multiple-choice options. It must rely entirely on its pre-trained internal knowledge to answer, without access to any additional resources or clarifications. This setting evaluates the model's intrinsic understanding of crystallographic principles and its ability to perform domain-specific reasoning based solely on its learned representations.
In contrast, the open-book mode supplements the question and answer choices with a brief, domain-relevant textual passage. This supporting material contains essential background information or clarifications pertinent to the question, similar to consulting a short textbook excerpt. Because this passage is fixed and expert-reviewed (not retrieved), open-book results estimate the generator-side ceiling under perfect retrieval (an oracle / Gold-Standard RAG scenario) rather than the performance of a full retriever+generator pipeline. The aim is to evaluate the model's ability to integrate external context with its internal knowledge in order to reason more effectively.
Together, these two evaluation settings provide complementary insights. While the closed-book mode tests baseline domain knowledge and internal consistency, the open-book mode examines the model's capacity to synthesize and apply external scientific information. This dual-mode structure is particularly valuable for identifying the limits of a model's independent expertise versus its ability to benefit from curated guidance.
\begin{figure}[ht]
\centering
\begin{tikzpicture}[
  font=\small,
  node distance=2.1cm,
  >=latex,
  align=center,
  box/.style={draw=black,rounded corners,fill=gray!5,inner sep=8pt,line width=0.8pt},
  wavy/.style={decorate, decoration={coil,aspect=0.1,segment length=3pt,amplitude=1.5pt}},
  curvy/.style={decorate, decoration={snake,amplitude=3pt,segment length=8pt}}
]
\node[box, rotate=90] (prompt) at (-3.3,0) {
  \begin{minipage}{0.22\textwidth}
    \centering
    \textbf{Question + Options}\\[2pt]
    \textit{(Closed-Book Input)}
  \end{minipage}
};
\node[box, rotate=-90] (support) at (3.3,0) {
  \begin{minipage}{0.2\textwidth}
    \centering
    \textbf{Supporting Textual Material}\\[2pt]
    \textit{(Open‑Book, GPT-4.5 Supporting Knowledge)}
  \end{minipage}
};
\node[box] (fusion) at (0,2.2) {
  \begin{minipage}{0.16\textwidth}
    \centering
    \textbf{Concatenate / Fuse}\\[2pt]
    \textit{(Combined Context)}
  \end{minipage}
};
\node[box] (llm) at (0,0) {
  \begin{minipage}{0.16\textwidth}
    \centering
    \textbf{LLM Inference}\\[2pt]
    \textit{(Generates Answer)}
  \end{minipage}
};
\node[box] (output) at (0,-2.2) {
  \begin{minipage}{0.12\textwidth}
    \centering
    \textbf{QA Result}\\[2pt]
    \textit{(Final Output)}
  \end{minipage}
};
\draw[->, line width=0.8pt, black] (prompt.east) to[out=30,in=150] (fusion.west);
\draw[->, line width=0.8pt, black] (support.west) to[out=150,in=30] (fusion.east);
\draw[->, line width=0.8pt, black] (fusion) -- (llm);
\draw[->, line width=0.8pt, black] (llm) -- (output);
\end{tikzpicture}
\caption{Illustration of our open-book QA pipeline for crystallography. 
In \textbf{closed-book} mode, the model sees only the question (left rotated box). 
In \textbf{open-book} mode, it also receives domain-specific supporting textual material (right rotated box), which is concatenated and fed to the LLM (center pipeline), producing the final QA result.}
\label{fig:openbookpipeline}
\end{figure}

\subsection{Supporting Textual Material Generation}
\subsubsection{AI-Generated Supporting Materials (Open‑Book, Without Expert Review)}
Due to copyright constraints and the impracticality of systematically scanning large volumes of actual textbooks, GPT-4.5 was used to generate short, supportive textual paragraphs for the open-book evaluations.

The generation process followed three key principles. First, each passage summarized fundamental crystallographic concepts relevant to the specific question. Second, care was taken to avoid directly stating the correct answer, ensuring that models must still reason rather than extract solutions verbatim. Third, the length and content were controlled to maintain clarity and focus, typically resulting in brief paragraphs spanning half to one page.
Our standardized generation protocol was: Given each question and answer choices (without revealing which option is correct), we prompted GPT-4.5 (temperature=0.7, top\_p=0.95, max\_tokens=800) with: “Given the following question: [QUESTION] and answer choices: [OPTIONS], provide a concise explanation of [RELEVANT CONCEPT] that would help answer this question without revealing the correct answer. Focus on fundamental principles, key definitions, and relevant physical mechanisms. Target length: 500–600 tokens.” Passages that revealed answers or contained technical errors were regenerated with revised prompts emphasizing guidance without revelation.

\subsubsection{AI-Generated Materials with Expert Review}
We engaged three PhD students specializing in crystallography to review and refine the supporting materials. These experts, with four to seven years of research experience in XRD, crystal structure analysis, and materials characterization, were instructed to correct any technical inaccuracies in AI-generated materials, improve explanations with precise domain terminology, ensure comprehensive coverage of relevant concepts, improve clarity while maintaining concision, remove potentially misleading information, and add critical contextual details missing from the original materials.

As demonstrated in Figure \ref{fig:expert-reviewed}, the expert-reviewed version provides substantial improvements over the initial AI-generated explanations. Furthermore, Figure \ref{fig:qa_example} illustrates a representative example from the multiple-choice question-answering dataset, showcasing the original question, corresponding answer choices, and the AI-generated explanation as revised by an expert.
\subsubsection{Quality Dimensions for Expert Review}
To make expert refinement reproducible, reviewers applied four explicit quality dimensions to all 217 passages: (1) \textbf{Technical accuracy}—terminology, equations, and physical principles must be correct and precise; (2) \textbf{Conceptual precision}—the targeted mechanism must be distinguished from related but distinct phenomena to avoid distraction-based errors; (3) \textbf{Pedagogical structure}—content is organized from definitions to mechanisms to question-specific implications; (4) \textbf{Contextual relevance}—remove tangential information and focus on the knowledge gap the question targets. Reviewers kept token counts within 5\% of the AI-generated versions to preserve token-matched comparisons.
\subsection{Input Context Structuring}
We implement a structured input preparation approach via a fusion module. For open-book evaluations, we employ a retrieve-then-read style input format (without executing retrieval) where the supporting material precedes the question:
\begin{equation}
x = \textsc{Format}(x_s, x_q, x_o)
\end{equation}
\noindent where $x_s$ is the supporting textual content, $x_q$ represents the question text, and $x_o$ indicates the provided answer choices. The \textsc{Format} function organizes these components with clear delimiters and instructional guidance. For closed-book evaluations, the supporting material is omitted ($x_s = \emptyset$), while the remaining formatting is preserved.
This structured approach aligns with established retrieval-augmented generation paradigms, wherein external knowledge serves as contextual reference material. Here, $x_s$ is an oracle helper passage selected and refined offline; no retriever, corpus chunking, or ranking is used in OPENXRD. Figure~\ref{fig:openbookpipeline} demonstrates the architecture of this open-book QA pipeline. Currently, our implementation is exclusively textual; however, this framework can be extended to multimodal data (such as XRD patterns and crystallographic diagrams) as generative vision models become sufficiently advanced.

\subsection{Evaluation Metrics}
We use several metrics to quantitatively assess model performance across evaluation modes, primarily focusing on accuracy, calculated as:
\begin{equation}
\text{Accuracy} = \frac{|\{\text{correctly answered questions}\}|}{|\{\text{all questions}\}|} \times 100\%
\end{equation}
For our 217-question benchmark, accuracy is reported both as an aggregated measure across the entire dataset and as a disaggregated measure by specific subtasks. For each model $\bm{M}$ and subtask $t \in \mathcal{T}$, accuracy is computed as:
\begin{equation}
\text{Accuracy}_{\bm{M},t} = \frac{|\{q \in Q_t : \bm{M}(q) = y_q\}|}{|Q_t|} \times 100\%
\end{equation}
\noindent where $Q_t$ is the set of questions associated with subtask $t$, $y_q$ is the correct answer, and $\bm{M}(q)$ denotes the model's prediction.
To assess the efficacy of open-book augmentation, we measure performance improvement $\Delta$ as:
\begin{equation}
\Delta_{\bm{M}} = \text{Accuracy}_{\bm{M}}^{\text{Open-Book}} - \text{Accuracy}_{\bm{M}}^{\text{Closed-Book}}
\label{eq:delta}
\end{equation}
Additionally, we quantify the relative improvement gained through expert-reviewed materials compared to AI-generated supporting materials:
\begin{equation}
\Delta_{\bm{M}}^{\text{Expert}} = \text{Accuracy}_{\bm{M}}^{\text{Open-Book, Expert}} - \text{Accuracy}_{\bm{M}}^{\text{Open-Book, AI}}
\end{equation}
This comprehensive evaluation framework allows us to explore several aspects. We examine the intrinsic crystallographic knowledge inherent in each model, referred to as closed-book performance. We also assess each model's ability to leverage external knowledge effectively, known as open-book improvement. Additionally, we evaluate the added value provided by expert-reviewed supporting materials, termed expert refinement differential. Furthermore, we analyze model-specific strengths and weaknesses across various crystallographic subtasks.

\subsection{Benchmark Reproducibility and Extension Guidelines}
\label{sec:repro_guidelines}

To establish OPENXRD as a reliable community benchmark with reproducible content standards, we provide documentation and tools for three primary use cases.

\textbf{Public release.} All benchmark materials are publicly available at \url{https://github.com/niaz60/OpenXRD} and will be archived on Zenodo with DOI upon acceptance. The release includes: full question set with answers and explanations; AI-generated and expert-reviewed supporting passages for all 217 questions; GPT-4.5 generation prompts with parameters; expert-review guidelines with the four-dimension quality rubric; token matching scripts; and validation protocols for answer-guidance testing.

\textbf{Evaluating new models.} Researchers should download the full benchmark, run the closed-book and open-book protocols (AI and expert passages) using our scripts, and compare against baseline tables to contextualize results.

\textbf{Creating new crystallography questions.} Follow our question template (prompt, 3–4 options, single correct answer, brief explanation, subtask label); generate initial passages with our GPT-4.5 prompt template (“Given question... provide a concise explanation... Target 500–600 tokens”); regenerate any passage that reveals the answer or has technical errors; apply expert review using the four dimensions; token-match AI and expert versions to <5\% difference ($r>0.90$).

\textbf{Domain adaptation.} For new scientific domains, adapt the generation prompt to domain terminology, recruit domain experts to review using the same four dimensions, validate answer-guidance properties on held-out models, and document domain-specific modifications.

\textbf{Validation protocol.} To ensure supporting materials are answer-guiding without being answer-revealing: (1) choose 3–5 held-out models across capacity tiers; (2) test on 20–30 questions in three conditions—closed-book, explicit-answer materials, candidate materials; (3) candidate materials pass if performance lies between closed-book and explicit-answer conditions and preserves the closed-book model ranking. This confirms that materials aid reasoning without providing shortcuts.
\subsection{Context Assimilation vs. Parameter Fine-Tuning}
\label{sec:scope_peft}

OPENXRD evaluates \emph{context assimilation}—how models utilize short, answer-guiding passages provided at inference time. We intentionally do not modify model weights through parameter-efficient fine-tuning (PEFT) methods such as LoRA~\cite{hu2021lora}, as this would target a fundamentally different research question: parametric knowledge embedding rather than inference-time guidance utilization.
Throughout this work, expert refinement refers to refinement of the supporting passages, not to any modification of evaluated LLMs. Accordingly, any performance changes in open-book mode should be interpreted as improved use of provided context (context assimilation), not weight-level refinement of model reasoning.

Several factors motivate this scope decision. First, many of the 74 models in our evaluation pool are API-only or closed-weight systems (GPT-4.5, O3-mini, Claude variants) that cannot be fine-tuned uniformly, making fair PEFT comparisons intractable. Second, our 217-question benchmark is intentionally reserved as a held-out evaluation set to preserve benchmark integrity; fine-tuning on these questions or near-duplicates would introduce data leakage and undermine reproducibility. Third, PEFT optimization conflates parametric adaptation effects with the inference-time assimilation dynamics we aim to isolate.

OPENXRD is complementary to PEFT research rather than competitive with it. Researchers can fine-tune open-weight models on external crystallographic corpora and subsequently use OPENXRD to test whether inference-time guidance remains beneficial, becomes redundant, or introduces interference after domain adaptation. This decomposition of parametric knowledge versus contextual reasoning is precisely the diagnostic capability OPENXRD provides. For researchers aiming to perform PEFT studies using our benchmark, we provide the complete dataset, evaluation scripts, and code framework on our project repository. This open-source release enables researchers to design their own PEFT experiments using standard techniques such as LoRA with appropriate data splits to prevent leakage between training and evaluation sets.

\section{Results}
\label{sec:experiments}

In this section, we present a detailed evaluation of our \methodName~ benchmark for crystallography, using a curated set of 217 XRD-related questions. Each question can be answered in two ways: Closed-book mode, where the model relies solely on its internal knowledge, and Open-book mode, where it consults supporting textual material. Our goal is to determine to what extent supporting textual material access bolsters model performance, particularly for advanced domain-specific queries. We conduct our experiments in two phases: first using AI-generated supporting materials provided by GPT-4.5, and then using expert-reviewed versions of these materials to assess whether human domain expertise further enhances performance. 

\subsection{Setup and Baselines}
\subsubsection{Models Compared}

We conduct a comprehensive evaluation across 74 state-of-the-art language models and vision-language models spanning multiple architectural families, parameter scales, and specialization domains. This extensive model selection enables systematic analysis of how model architecture, scale, and domain adaptation affect crystallographic reasoning performance under both closed-book and open-book conditions.

\textit{OpenAI models.} We evaluate 13 OpenAI models across multiple generations and capabilities. The next-generation GPT-5 series includes \texttt{gpt-5} and \texttt{gpt-5-codex}\cite{openai2025gpt5,openai2025gpt5codex}. The reasoning-optimized O-series comprises \texttt{o3-mini} (o3-mini-2025-01-31), \texttt{o1} (o1-2024-12-17), and \texttt{o1-mini}\cite{openai2025o3mini,openai2024o1systemcard}. The GPT-4 family includes \texttt{gpt-4.5-preview} (gpt-4.5-preview-2025-02-27), \texttt{gpt-4o}, \texttt{gpt-4o-mini}, \texttt{gpt-4-turbo} (gpt-4-turbo-2024-04-09), \texttt{gpt-4-turbo-preview} (gpt-4-preview-0125), \texttt{gpt-4-0314} (gpt-4-0613), and the base \texttt{gpt-4}\cite{openai2025gpt45,openai2024gpt4o,openai2023gpt4turbo,openai2023gpt4}. We also evaluate earlier-generation models \texttt{gpt-3.5-turbo} and \texttt{gpt-3.5-turbo-16k}\cite{openai2023gpt35turbo,openai2023gpt35turbo16k}.
\textit{Anthropic Claude models.} We assess four Claude variants spanning two generations: \texttt{claude-3.5-sonnet}, \texttt{claude-3.5-haiku}, \texttt{claude-3-opus}, and \texttt{claude-3-haiku}\cite{anthropic2024claude3,anthropic2024claude35sonnet,anthropic2024claude35haiku}.
\textit{Meta LLaMA models.} The LLaMA family is represented by six models across three generations and multiple parameter scales: \texttt{llama-3.1-405b-instruct} (405B parameters), \texttt{llama-3.1-70b-instruct} and \texttt{llama-3-70b-instruct} (70B parameters), \texttt{llama-3.1-8b-instruct} and \texttt{llama-3-8b-instruct} (8B parameters), and \texttt{llama-3.2-3b-instruct} (3B parameters)\cite{dubey2024llama3herd,metallamateam2024llama32}.

\textit{QWEN/Alibaba models.} We evaluate 10 models from Alibaba's QWEN ecosystem. General-purpose models include \texttt{qwen3-next-80b-a3b-instruct}, \texttt{qwen3-next-80b-a3b-thinking} (reasoning-optimized variant), \texttt{qwen-2.5-72b-instruct}, \texttt{qwen-plus}, \texttt{qwen-max}, and \texttt{qwen-2.5-7b-instruct}\cite{qwen3,qwen25}. Code-specialized variants include \texttt{qwen3-coder-flash}, \texttt{qwen3-coder-plus}, and \texttt{qwen-2.5-coder-32b-instruct}\cite{qwen3coder,hui2024qwen25coder}. We also evaluate \texttt{tongyi-deepresearch-30b-a3b}, Alibaba's research-focused model\cite{tongyideepresearch}.
\textit{Mistral AI models.} Seven Mistral variants are evaluated: \texttt{mistral-large} and \texttt{mistral-small} (dense models)\cite{mistral2024large,mistral2024small}, \texttt{mistral-7b-instruct} and \texttt{mistral-7b-instruct-v0.1} (7B base models)\cite{jiang2023mistral}, the mixture-of-experts models \texttt{mixtral-8x22b-instruct} and \texttt{mixtral-8x7b-instruct}\cite{jiang2024mixtral,mistral2024mixtral8x22b}, and the multimodal \texttt{pixtral-12b}\cite{agrawal2024pixtral}. \textit{DeepSeek models.} We evaluate two DeepSeek variants: \texttt{deepseek-v3.1-terminus} and \texttt{deepseek-chat}\cite{deepseekv3}.
\textit{Google models.} Three Google models are assessed: the flagship \texttt{gemini-2.5-pro} and the open-weight models \texttt{gemma-2-27b-it} and \texttt{gemma-2-9b-it}\cite{gemini2025v2_5,gemma2v2}.
\textit{Amazon Nova models.} We evaluate Amazon's Nova series across three capability tiers: \texttt{nova-pro-v1}, \texttt{nova-lite-v1}, and \texttt{nova-micro-v1}\cite{AmazonAGI2024Nova}.
\textit{X.AI models.} We include X.AI's \texttt{grok-4-fast} in our evaluation\cite{xAI2025Grok4Fast}.
\textit{LLaVA vision-language models.} Ten LLaVA variants are evaluated, spanning multiple generations and backbone architectures. These include \texttt{llava-v1.6-34b} and \texttt{llava-v1.6-mistral-7b} from the 1.6 series\cite{liu2024llavanext}, \texttt{llava-v1.5-13b} and \texttt{llava-v1.5-7b} from the 1.5 series\cite{liu2023improvedllava}, and six LLaVA-OneVision models with QWEN2 backbones: \texttt{llava-onevision-qwen2-7b-si}, \texttt{llava-onevision-qwen2-7b-ov-chat}, \texttt{llava-onevision-qwen2-7b-ov} (7B variants), and \texttt{llava-onevision-qwen2-0.5b-ov}, \texttt{llava-onevision-qwen2-0.5b-si} (0.5B variants)\cite{li2024llava-onevision}. These models combine vision encoders (ViT) with LLM backbones (Mistral, LLaMA, QWEN) for multimodal understanding. In our text-only evaluation setting, these models process supporting textual materials that may include image captions but not the actual images themselves.
\textit{LLaMAT crystallography-specialized models:} We evaluate three domain-adapted LLaMAT variants that have been pre-trained on crystallographic information files (CIF): \texttt{llamat-3-chat}, \texttt{llamat-2-chat}, and \texttt{llamat-2}\cite{mishra2024llamat}. These models represent explicit attempts to enhance materials science reasoning through domain-specific pre-training.

\textit{Dziner models.} We assess three Dziner-QWEN variants spanning different parameter scales: \texttt{dziner-qwen-2.5-72b}, \texttt{dziner-qwen-2.5-coder-32b}, and \texttt{dziner-qwen-2.5-7b}\cite{ansari2024dziner}.

\textit{Perplexity models.} Two Perplexity variants are evaluated: \texttt{sonar-pro} and \texttt{sonar}\cite{perplexity2025sonarpro,perplexity2025sonar_new}.
\textit{Additional specialized models.} Our evaluation includes several community-developed and specialized models: Meituan's \texttt{longcat-flash-chat}\cite{meituan_longcat_flash_2025}, Arcee-AI's \texttt{afm-4.5b}\cite{arcee_afm_2025}, the community-tuned models \texttt{undi95/remm-slerp-l2-13b} and \texttt{gryphe/mythomax-l2-13b}\cite{undi95_remm_slerp_2023,gryphe_mythomax_2023}, the vision-language models \texttt{honeybee-13b} and \texttt{honeybee-7b}\cite{cha2023honeybee_arxiv}, and the routing system \texttt{openrouter/auto}\cite{openrouter_auto_2023}.

This comprehensive model selection, spanning from 0.5B to 405B+ parameters and covering generalist, reasoning-optimized, code-specialized, domain-adapted, and vision-language architectures, enables robust analysis of how different model capabilities interact with domain-specific supporting materials in crystallographic question answering.

Several models were considered but ultimately excluded from our evaluation due to architectural incompatibilities or technical limitations. The \texttt{Ether-0} model\cite{narayanan2025training}, designed specifically as a reward model for scoring the quality of scientific reasoning in chemistry rather than as a generative question-answering system, was deemed inappropriate for our benchmark. Reward models are trained using preference learning objectives to rank or score candidate responses, fundamentally differing from generative models that produce answers directly. This architectural mismatch makes \texttt{Ether-0} incompatible with our multiple-choice QA evaluation framework, which requires models to generate or select answers rather than evaluate them.

We also encountered technical challenges with several LLaMAT variants—specialized adaptations of LLaMA models pre-trained on crystallographic information files (CIF). While three LLaMAT models (\texttt{llamat-3-chat}, \texttt{llamat-2-chat}, and \texttt{llamat-2}) were successfully evaluated after configuring \texttt{max\_model\_len=2048} to accommodate their tokenization schemes, three additional variants (\texttt{llamat-3-base}, \texttt{llamat-3-cif}, and \texttt{llamat-2-cif}) exhibited incomplete HuggingFace configurations: the base \texttt{llamat-3} model lacked proper architecture specifications in \texttt{config.json}, while both CIF-specialized variants had malformed model cards with missing tokenizer bindings that prevented inference initialization. Given that the successfully evaluated chat-optimized LLaMAT variants achieved moderate baseline performance (50.69-57.14\% accuracy in closed-book mode; see Table~\ref{tab:closedbook-all}), the exclusion of these broken variants does not materially impact our conclusions about crystallography-specialized models.

To facilitate systematic performance analysis across different capability levels, we categorize the 74 evaluated models into three size groups based on parameter count and architectural complexity, as summarized in Table~\ref{tab:model_categories}. This categorization enables structured comparison of how model scale affects the ability to assimilate external domain knowledge in crystallography tasks.

\begin{table}[H]
\centering
\caption{Model categorization by size for systematic performance analysis. The 74 evaluated models are grouped into three categories to examine scale-dependent effects on external knowledge assimilation in crystallography.}
\label{tab:model_categories}
\resizebox{\columnwidth}{!}{%
\begin{tabular}{lll}
\toprule
\textbf{Category} & \textbf{Size Range} & \textbf{Representative Models} \\
\midrule
Small models & $<$ 10B parameters & Mistral-7B, Phi-3-Mini (3.8B), Llama-3-8B, \\
 & & LLaVA-v1.5-7B, QWEN2-0.5B, HoneyBee-7B, \\
 & & LLaMAT-2 (7B), AFM-4.5B \\
\midrule
Mid-sized models & 10B--70B parameters & Llama-3-70B, LLaVA-v1.6-34B, QWEN-2.5-32B, \\
 & & Mixtral-8$\times$7B ($\sim$56B active), \\
 & & Gemma-2-27B, Mistral-Small (24B) \\
\midrule
Large models & $>$ 70B parameters & Llama-3.1-405B, QWEN-2.5-72B, \\
 & (or closed-weight & dziner-qwen-2.5-72B, GPT-4, GPT-5, \\
 & API models) & GPT-4.5, O3-mini, Claude-3-Opus, \\
 & & Gemini-2.5-Pro, DeepSeek-V3.1 \\
\bottomrule
\end{tabular}%
}
\end{table}

\subsubsection{Inference Setup}
We employ a zero-shot inference paradigm for all models without parameter updating or fine-tuning on our crystallography evaluation set. For each query, we construct a prompt consisting of the question and enumerated options, formatted consistently across all models. No in-context learning examples or demonstrations are provided, requiring models to rely entirely on their parametric knowledge. All foundation models are evaluated using their published weights and architectures; O-family models have documented parameter-efficient training on scientific corpora, while vision-language models in the LLaVA family were primarily trained on general domain text-image pairs without domain-specific adaptation to crystallography.

\subsubsection{Experimental Progression}
Our experimental approach follows two distinct phases: In the first phase, we use GPT-4.5 to generate supporting textual materials for each question, carefully prompting it to provide relevant information without revealing the answer. Following this, in the second phase, we engaged three Ph.D. students specializing in crystallography, with four to seven years of research experience, to review and refine the supporting materials for improved accuracy and pedagogical clarity. This two-phase approach allows us to quantify the value added by human domain expertise in the construction of supporting materials relative to state-of-the-art AI generation.

\subsection{Closed-Book Mode Observations} 
\begin{figure}[htbp]
    \centering
    \includegraphics[width=\linewidth]{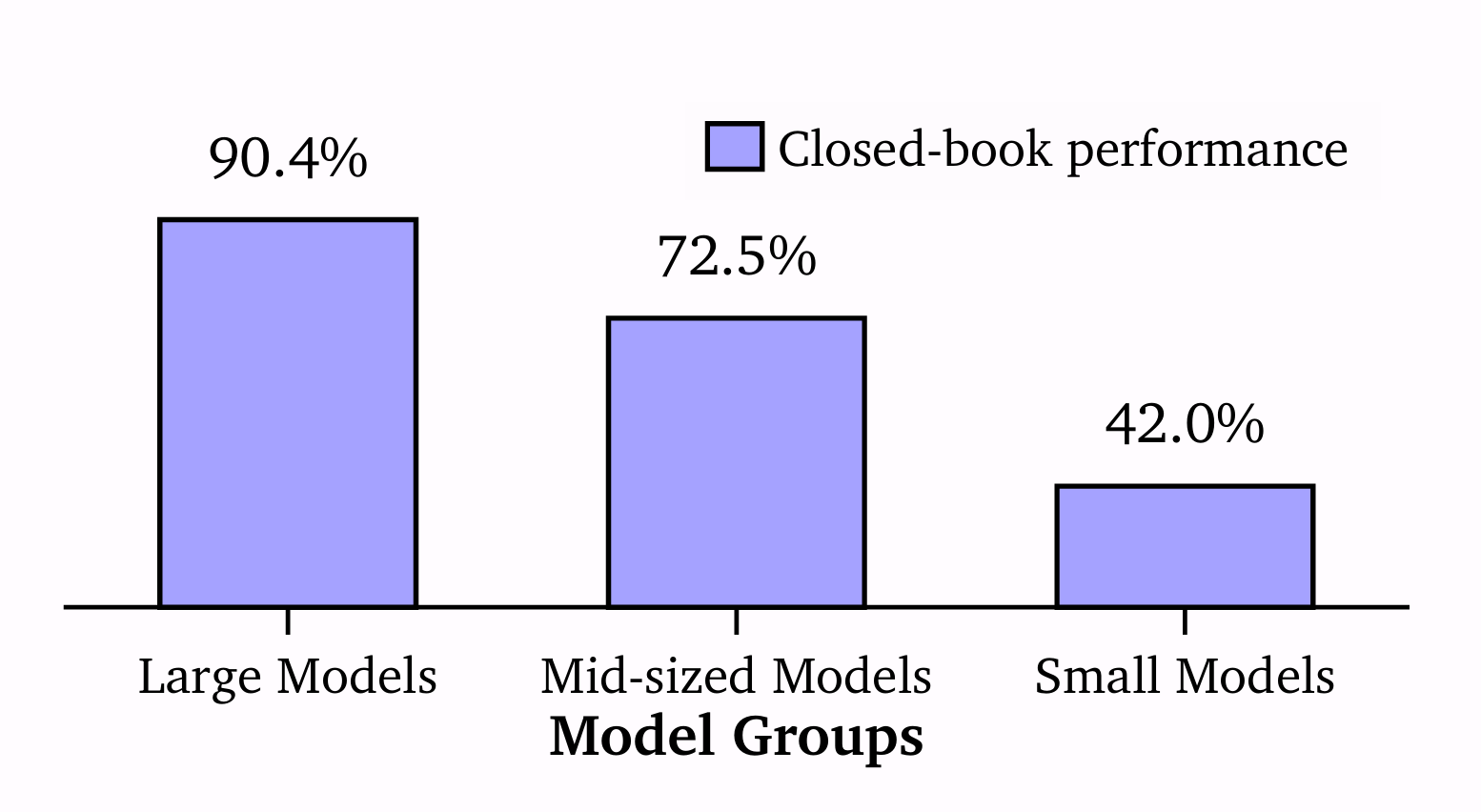}
    \caption{Comparison of closed-book performance by model size group. Large Models ($>$70B parameters or advanced architectures like GPT-4, GPT-5, O1, O3-mini, n=27); Mid-sized Models ($>$7B-70B parameters including LLaVA-34B, QWEN2-7B, Mistral-7B, various 8B-13B models, n=22); Small Models ($\leq$7B parameters including QWEN2-0.5B, LLaMA-3.2-3B, Honeybee-7B, LLaMAT-2, n=25). Analysis excludes unreliable llava-v1.6-vicuna variants.}
    \label{fig:model_size_performance}
\end{figure}

Table~\ref{tab:closedbook-all} presents the closed-book mode accuracy achieved on our 217-question crystallography benchmark, and Figure \ref{fig:model_size_performance} illustrates the categorized closed-book performance of the models.

O3-mini performs extremely well, achieving about 93.55\% accuracy, even surpassing some GPT-4 variants, possibly reflecting training or optimization details not disclosed; we report the observed performance. However, smaller LLaVA-based models face challenges with advanced reflection extinctions, often incorrectly identifying which plane indices vanish for BCC or confusing zone-axis notation. Despite these issues, nearly all models correctly answer basic factual questions like the number of crystal systems and who discovered X-ray diffraction.

\begin{table}[t]
\centering
\caption{Comprehensive "closed-book mode" accuracy results on the 217-item crystallography QA benchmark, including older GPT‑4 variants and newly reported models. The rightmost column shows model parameter sizes: B = billions, T = trillions, U = undisclosed. For Mixture-of-Experts (MoE) models, format is total/active parameters. Estimated values marked with $\sim$, N/A for routing services.}
\label{tab:closedbook-all}
\resizebox*{\columnwidth}{0.9\textheight}{%
\begin{tabular}{r l c c r}
\toprule
\textbf{Rank} & \textbf{Model (Closed-book mode)} & \textbf{Acc. (\%)} & \textbf{Correct} & \textbf{Params} \\
\midrule

1 & openai/gpt-5 & 96.77 & 210/217 & $\sim$300B \\
2 & x-ai/grok-4-fast & 96.31 & 209/217 & U \\
2 & openai/gpt-5-codex & 96.31 & 209/217 & $\sim$300B \\
4 & google/gemini-2.5-pro & 95.39 & 207/217 & U \\
5 & o3-mini & 93.55 & 203/217 & U \\
6 & meituan/longcat-flash-chat & 93.09 & 202/217 & 560B/27B \\
7 & gpt-4.5-preview & 92.63 & 201/217 & U \\
8 & anthropic/claude-3.5-sonnet & 91.24 & 198/217 & $\sim$175B \\
9 & openai/gpt-4o & 90.74 & 196/216 & $\sim$200B \\
10 & dziner-qwen-2.5-72b & 90.32 & 196/217 & $\sim$72B \\
10 & perplexity/sonar-pro & 90.32 & 196/217 & 70B \\
10 & qwen/qwen3-next-80b-a3b-instruct & 90.32 & 196/217 & 80B/3B \\
13 & deepseek/deepseek-v3.1-terminus & 89.86 & 195/217 & 671B/37B \\
13 & qwen/qwen-plus & 89.86 & 195/217 & U \\
13 & qwen/qwen3-next-80b-a3b-thinking & 89.86 & 195/217 & 80B/3B \\
16 & deepseek/deepseek-chat & 89.40 & 194/217 & 671B/37B \\
16 & o1 & 89.40 & 194/217 & $\sim$200B \\
16 & qwen/qwen-max & 89.40 & 194/217 & U \\
19 & anthropic/claude-3-opus & 88.94 & 193/217 & U \\
20 & openai/o1-mini & 88.02 & 191/217 & $\sim$100B \\
21 & meta-llama/llama-3.1-405b-instruct & 87.56 & 190/217 & 405B \\
21 & qwen/qwen-2.5-72b-instruct & 87.56 & 190/217 & 72B \\
23 & mistralai/mistral-large & 86.18 & 187/217 & 123B \\
24 & amazon/nova-pro-v1 & 86.11 & 186/216 & $\sim$90B \\
25 & gpt-4-turbo & 85.25 & 185/217 & 1.8T/280B \\
26 & meta-llama/llama-3-70b-instruct & 84.79 & 184/217 & 70B \\
26 & meta-llama/llama-3.1-70b-instruct & 84.79 & 184/217 & 70B \\
28 & dziner-qwen-2.5-coder-32b & 83.87 & 182/217 & $\sim$32B \\
29 & openai/gpt-4-0314 & 83.41 & 181/217 & 1.8T/280B \\
29 & gpt-4-turbo-preview & 83.41 & 181/217 & 1.8T/280B \\
31 & amazon/nova-lite-v1 & 82.03 & 178/217 & $\sim$20B \\
32 & dziner-qwen-2.5-7b & 81.57 & 177/217 & $\sim$7B \\
32 & gpt-4 & 81.57 & 177/217 & 1.8T/280B \\
34 & openai/gpt-4o-mini & 81.11 & 176/217 & $\sim$8B \\
34 & openrouter/auto & 81.11 & 176/217 & N/A \\
36 & qwen/qwen-2.5-7b-instruct & 79.72 & 173/217 & 7B \\
37 & google/gemma-2-27b-it & 79.26 & 172/217 & 27B \\
38 & anthropic/claude-3.5-haiku & 77.57 & 166/214 & U \\
39 & mistralai/mixtral-8x22b-instruct & 76.81 & 159/207 & 141B/39B \\
40 & mistralai/mistral-7b-instruct & 75.35 & 162/215 & 7B \\
41 & google/gemma-2-9b-it & 75.12 & 163/217 & 9B \\
42 & amazon/nova-micro-v1 & 74.65 & 162/217 & $\sim$11B \\
42 & mistralai/mistral-small & 74.65 & 162/217 & 24B \\
44 & anthropic/claude-3-haiku & 74.04 & 154/208 & $\sim$20B \\
45 & openai/gpt-3.5-turbo-16k & 72.60 & 151/208 & $\sim$20B \\
46 & mistralai/mixtral-8x7b-instruct & 72.33 & 149/206 & 47B/13B \\
47 & meta-llama/llama-3-8b-instruct & 71.89 & 156/217 & 8B \\
48 & mistralai/pixtral-12b & 71.83 & 153/213 & 12B \\
49 & openai/gpt-3.5-turbo & 70.67 & 147/208 & $\sim$20B \\
50 & meta-llama/llama-3.1-8b-instruct & 69.12 & 150/217 & 8B \\
51 & llava-v1.6-34b & 66.82 & 145/217 & 34B \\
52 & lmms-lab/llava-onevision-qwen2-7b-si & 66.36 & 144/217 & 7B \\
53 & lmms-lab/llava-onevision-qwen2-7b-ov-chat & 65.90 & 143/217 & 7B \\
54 & lmms-lab/llava-onevision-qwen2-7b-ov & 65.44 & 142/217 & 7B \\
55 & meta-llama/llama-3.2-3b-instruct & 64.98 & 141/217 & 3B \\
56 & arcee-ai/afm-4.5b & 62.21 & 135/217 & 4.5B \\
57 & mistralai/mistral-7b-instruct-v0.1 & 59.50 & 119/200 & 7B \\
58 & perplexity/sonar & 59.24 & 125/211 & 70B \\
59 & undi95/remm-slerp-l2-13b & 57.08 & 121/212 & 13B \\
60 & llamat-3-chat & 57.14 & 124/217 & 8B \\
61 & alibaba/tongyi-deepresearch-30b-a3b & 55.76 & 121/217 & 30B/3B \\
62 & gryphe/mythomax-l2-13b & 53.77 & 114/212 & 13B \\
63 & llava-v1.6-mistral-7b & 52.99 & 115/217 & 7B \\
64 & llamat-2-chat & 50.69 & 110/217 & 7B \\
65 & lmms-lab/llava-onevision-qwen2-0.5b-ov & 47.47 & 103/217 & 0.5B \\
66 & llava-v1.5-13b & 46.54 & 101/217 & 13B \\
67 & lmms-lab/llava-onevision-qwen2-0.5b-si & 46.08 & 100/217 & 0.5B \\
68 & qwen/qwen3-coder-flash & 43.46 & 83/191 & U \\
69 & qwen/qwen3-coder-plus & 22.99 & 43/187 & U \\
70 & honeybee-13b & 22.12 & 48/217 & 13B \\
71 & qwen/qwen-2.5-coder-32b-instruct & 21.03 & 45/214 & 32B \\
72 & honeybee-7b & 19.35 & 42/217 & 7B \\
73 & llava-v1.5-7b & 17.97 & 39/217 & 7B \\
74 & llamat-2 & 16.59 & 36/217 & 7B \\
\bottomrule
\end{tabular}%
}
\end{table}

\subsection{Open-Book Mode Observations Using AI-Generated Materials (No Expert Review)}
\label{sec:openbook-ai}
Table~\ref{tab:openbook-ai-results} lists the closed-book mode vs. open-book mode accuracy on our 217-question crystallography benchmark. Here and in subsequent tables, $\Delta$ represents the performance improvement defined in Equation~\eqref{eq:delta}, with model subscripts omitted as each row corresponds to a different model. We first evaluate open-book mode accuracy by giving each model relevant supporting textual material generated by GPT-4.5.

We observe distinct performance patterns across model scales, architectures, and domain specializations. Remarkably, specialized coder models demonstrate the most dramatic improvements: qwen3-coder-plus gains +65.95\% and qwen3-coder-flash gains +40.87\% from AI-generated supporting materials, suggesting these models have significant reasoning capacity but minimal crystallographic knowledge. Among general-purpose models, mid-sized architectures (>7B–70B parameters) show substantial benefits, with models like llama-3.1-8b-instruct (+6.00\%), llava-onevision-qwen2-7b-ov-chat (+6.45\%), anthropic/claude-3-haiku (+8.45\%), and llava-v1.6-34b (+5.99\%) gaining meaningfully from domain-specific context. Larger models (70B+) such as llama-3.1-405b-instruct (-4.61\%), llama-3-70b-instruct (-4.14\%), and qwen-2.5-72b-instruct (-1.38\%) show minimal improvement or slight degradation, while very large frontier models like o3-mini see only minor gains (-4.15\%), suggesting they already possess substantial internal crystallographic knowledge.

Notably, multiple frontier models show performance degradation with AI-generated supporting materials: GPT-4.5-preview (-2.31\%), Claude-3.5-Sonnet (-1.84\%), dziner-qwen-2.5-72b (-3.22\%), and mistral-large (-2.31\%). This counterintuitive pattern reveals an important failure mode: for models with comprehensive pre-trained knowledge, additional context—even when domain-relevant—can introduce interference rather than assistance.

\begin{figure}[ht]
   \centering
   \includegraphics[width=\linewidth]{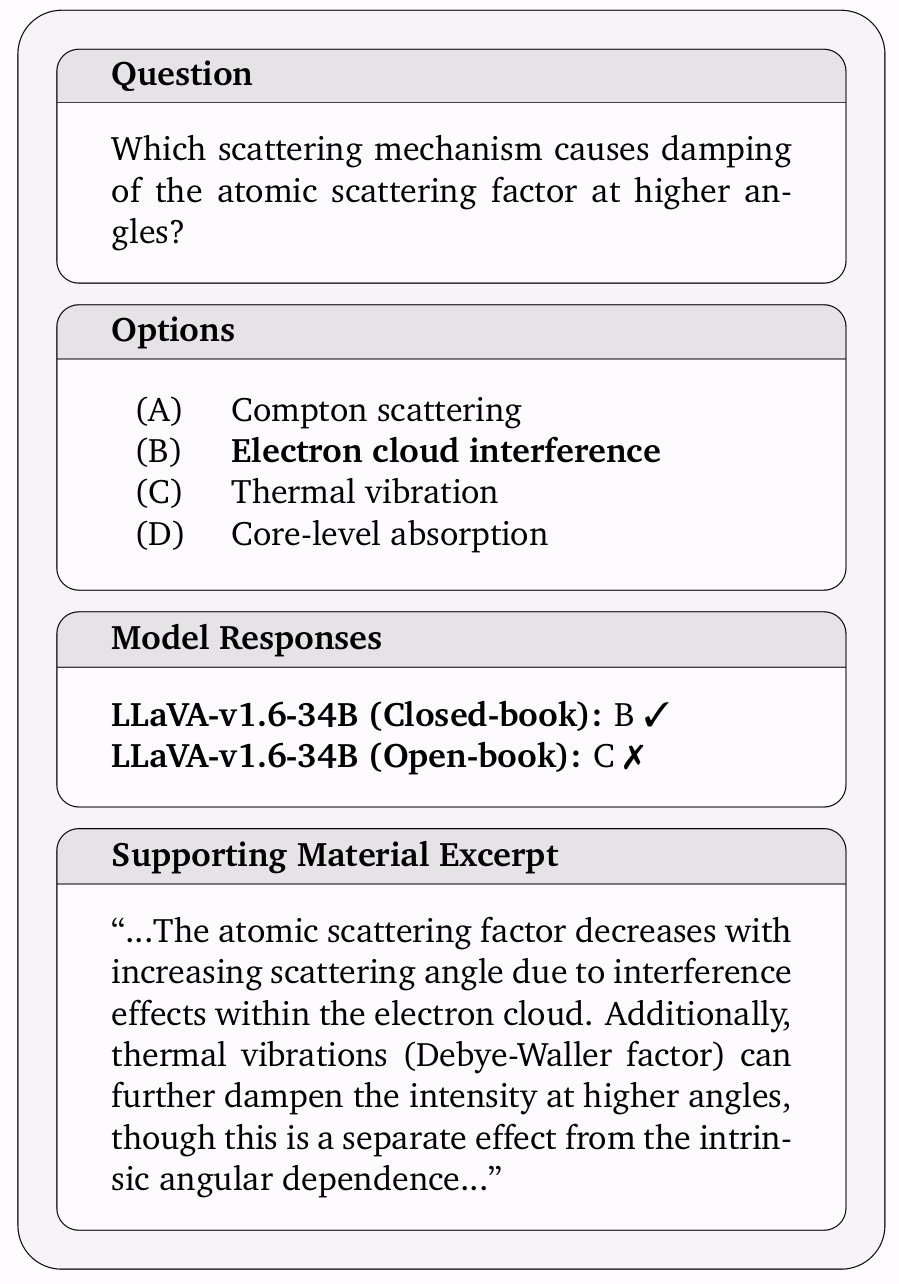}
   \caption{Example of confusion in LLaVA-v1.6-34B due to conflicting supporting material. The model had correct knowledge in closed-book mode but was misled by additional content that mentioned both the correct answer and a distractor as possible factors affecting scattering, albeit in different ways.}
   \label{fig:error_case}
\end{figure}

Figure~\ref{fig:error_case} shows a concrete example where supporting materials have caused confusion, directly illustrating this interference mechanism. In this case, LLaVA-v1.6-34B answered correctly in closed-book mode (Option B: Electron cloud interference) but was misled in open-book mode to select Option C (Thermal vibration). The supporting material discussed both phenomena in close proximity: ``The atomic scattering factor decreases with increasing scattering angle due to interference effects within the electron cloud. Additionally, thermal vibrations (Debye-Waller factor) can further dampen the intensity...'' This juxtaposition caused the model to conflate two distinct mechanisms—the intrinsic angular dependence (correct answer) versus temperature-dependent effects (distractor)—demonstrating that the failure mode is primarily \textit{conceptual interference} and \textit{distraction from related concepts} rather than simple information overload or redundancy.

\begin{figure}[htbp]
    \centering
    \includegraphics[width=\linewidth]{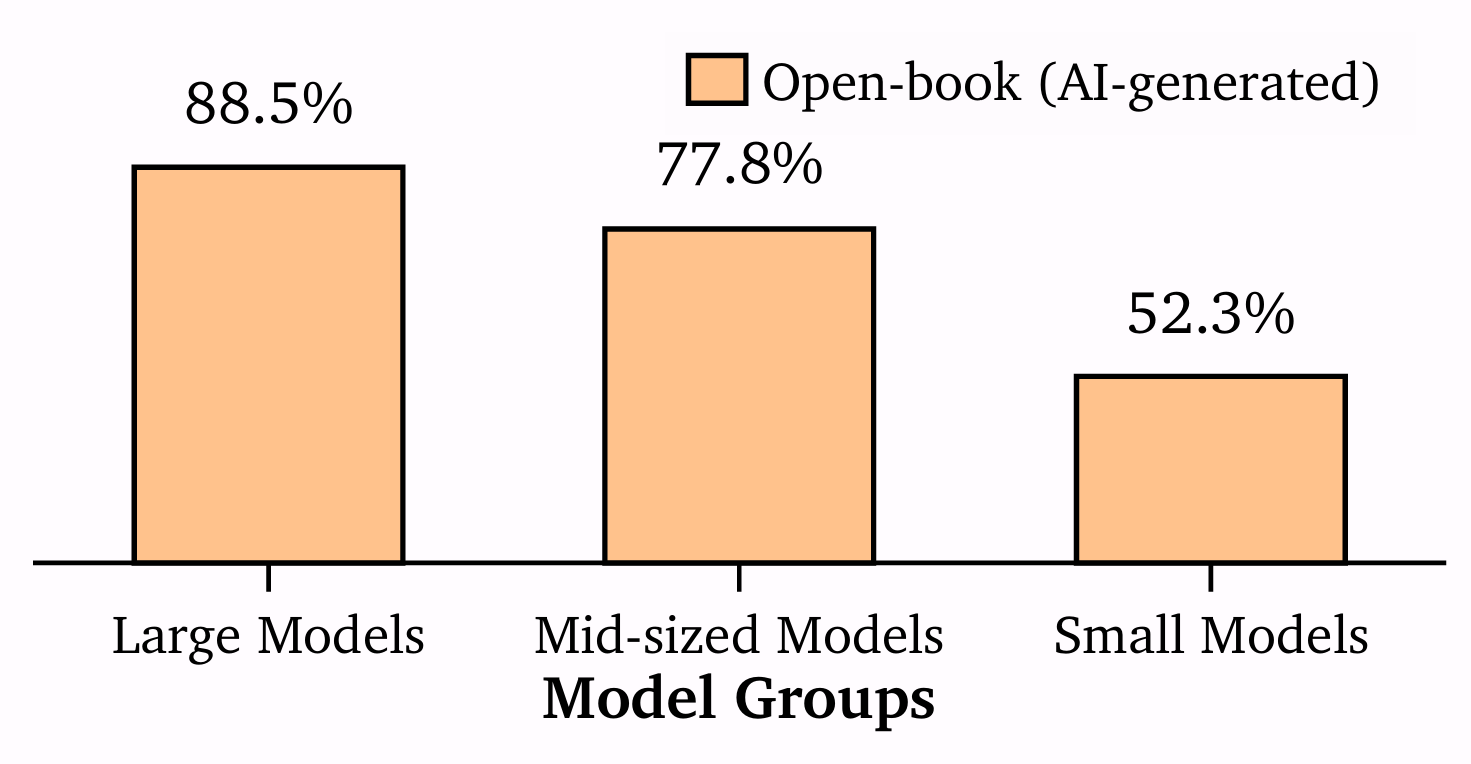}
    \caption{Open-book performance with AI-generated supporting materials by model size group. Large Models ($>$70B parameters or advanced architectures, n=27); Mid-sized Models ($>$7B-70B parameters, n=22); Small Models ($\leq$7B parameters, n=25). Compared to closed-book performance (Figure~\ref{fig:model_size_performance}), large models show slight degradation ($-1.9\%$), while mid-sized models gain $+5.3\%$ and small models gain $+10.3\%$, demonstrating that AI-generated context benefits models with knowledge gaps but can interfere with already-capable models.}
    \label{fig:openbook_ai_performance}
\end{figure}

Our detailed analysis of the AI-generated supporting materials revealed several limitations: occasional technical inaccuracies in specialized crystallographic terminology, inadequate depth of explanation for particularly complex phenomena, information presented at times being too general to help with specific question nuances, and in some cases, the supporting materials were tangential to the precise knowledge needed for the question. Figure~\ref{fig:openbook_ai_performance} summarizes these patterns by model size group, showing that mid-sized and small models benefit substantially from AI-generated materials ($+5.3\%$ and $+10.3\%$ respectively), while large models experience slight degradation ($-1.9\%$).

These observations led us to hypothesize that expert-reviewed materials might yield greater performance improvements, particularly for mid-sized models with significant capacity but incomplete domain knowledge.

\begin{table}[t]
\centering
\caption{Comparison of model accuracy in closed-book mode vs.\ open-book mode with AI-generated supporting materials. $\Delta =$ (Open-book mode) $-$ (Closed-book mode).}
\label{tab:openbook-ai-results}

\resizebox{\columnwidth}{!}{%
\begin{tabular}{l c c c}
\toprule
\textbf{Model} & \textbf{Closed-book mode (\%)} & \textbf{Open-book mode (\%)} & $\Delta$ \\
\midrule
openai/gpt-5 & 96.77 & 94.93 & -1.84 \\
x-ai/grok-4-fast & 96.31 & 95.39 & -0.92 \\
openai/gpt-5-codex & 96.31 & 94.47 & -1.84 \\
google/gemini-2.5-pro & 95.39 & 88.48 & -6.91 \\
o3-mini & 93.55 & 89.40 & -4.15 \\
meituan/longcat-flash-chat & 93.09 & 90.32 & -2.77 \\
gpt-4.5-preview & 92.63 & 90.32 & -2.31 \\
anthropic/claude-3.5-sonnet & 91.24 & 89.40 & -1.84 \\
openai/gpt-4o & 90.74 & 87.56 & -3.18 \\
dziner-qwen-2.5-72b & 90.32 & 87.10 & -3.22 \\
perplexity/sonar-pro & 90.32 & 90.32 & +0.00 \\
qwen/qwen3-next-80b-a3b-instruct & 90.32 & 89.40 & -0.92 \\
deepseek/deepseek-v3.1-terminus & 89.86 & 90.78 & +0.92 \\
qwen/qwen-plus & 89.86 & 90.78 & +0.92 \\
qwen/qwen3-next-80b-a3b-thinking & 89.86 & 87.10 & -2.76 \\
deepseek/deepseek-chat & 89.40 & 91.24 & +1.84 \\
o1 & 89.40 & 88.02 & -1.38 \\
qwen/qwen-max & 89.40 & 88.02 & -1.38 \\
anthropic/claude-3-opus & 88.94 & 88.02 & -0.92 \\
openai/o1-mini & 88.02 & 88.48 & +0.46 \\
meta-llama/llama-3.1-405b-instruct & 87.56 & 82.95 & -4.61 \\
qwen/qwen-2.5-72b-instruct & 87.56 & 86.18 & -1.38 \\
mistralai/mistral-large & 86.18 & 83.87 & -2.31 \\
amazon/nova-pro-v1 & 86.11 & 86.64 & +0.53 \\
gpt-4-turbo & 85.25 & 85.71 & +0.46 \\
meta-llama/llama-3-70b-instruct & 84.79 & 80.65 & -4.14 \\
meta-llama/llama-3.1-70b-instruct & 84.79 & 83.41 & -1.38 \\
dziner-qwen-2.5-coder-32b & 83.87 & 84.79 & +0.92 \\
openai/gpt-4-0314 & 83.41 & 77.42 & -5.99 \\
gpt-4-turbo-preview & 83.41 & 82.49 & -0.92 \\
amazon/nova-lite-v1 & 82.03 & 84.79 & +2.76 \\
dziner-qwen-2.5-7b & 81.57 & 79.26 & -2.31 \\
gpt-4 & 81.57 & 82.03 & +0.46 \\
openai/gpt-4o-mini & 81.11 & 82.03 & +0.92 \\
openrouter/auto & 81.11 & 81.57 & +0.46 \\
qwen/qwen-2.5-7b-instruct & 79.72 & 75.12 & -4.60 \\
google/gemma-2-27b-it & 79.26 & 78.80 & -0.46 \\
anthropic/claude-3.5-haiku & 77.57 & 88.02 & +10.45 \\
mistralai/mixtral-8x22b-instruct & 76.81 & 80.09 & +3.28 \\
mistralai/mistral-7b-instruct & 75.35 & 77.42 & +2.07 \\
google/gemma-2-9b-it & 75.12 & 77.42 & +2.30 \\
amazon/nova-micro-v1 & 74.65 & 79.26 & +4.61 \\
mistralai/mistral-small & 74.65 & 79.26 & +4.61 \\
anthropic/claude-3-haiku & 74.04 & 82.49 & +8.45 \\
openai/gpt-3.5-turbo-16k & 72.60 & 71.01 & -1.59 \\
mistralai/mixtral-8x7b-instruct & 72.33 & 75.96 & +3.63 \\
meta-llama/llama-3-8b-instruct & 71.89 & 73.27 & +1.38 \\
mistralai/pixtral-12b & 71.83 & 75.58 & +3.75 \\
openai/gpt-3.5-turbo & 70.67 & 71.50 & +0.83 \\
meta-llama/llama-3.1-8b-instruct & 69.12 & 75.12 & +6.00 \\
llava-v1.6-34b & 66.82 & 72.81 & +5.99 \\
lmms-lab/llava-onevision-qwen2-7b-si & 66.36 & 71.89 & +5.53 \\
lmms-lab/llava-onevision-qwen2-7b-ov-chat & 65.90 & 72.35 & +6.45 \\
lmms-lab/llava-onevision-qwen2-7b-ov & 65.44 & 71.43 & +5.99 \\
meta-llama/llama-3.2-3b-instruct & 64.98 & 62.67 & -2.31 \\
arcee-ai/afm-4.5b & 62.21 & 64.52 & +2.31 \\
mistralai/mistral-7b-instruct-v0.1 & 59.50 & 54.03 & -5.47 \\
perplexity/sonar & 59.24 & 68.84 & +9.60 \\
undi95/remm-slerp-l2-13b & 57.08 & 61.50 & +4.42 \\
llamat-3-chat & 57.14 & 30.41 & -26.73 \\
alibaba/tongyi-deepresearch-30b-a3b & 55.76 & 80.18 & +24.42 \\
gryphe/mythomax-l2-13b & 53.77 & 61.68 & +7.91 \\
llava-v1.6-mistral-7b & 52.99 & 58.53 & +5.54 \\
llamat-2-chat & 50.69 & 21.13 & -29.56 \\
lmms-lab/llava-onevision-qwen2-0.5b-ov & 47.47 & 51.15 & +3.68 \\
llava-v1.5-13b & 46.54 & 44.24 & -2.30 \\
lmms-lab/llava-onevision-qwen2-0.5b-si & 46.08 & 51.15 & +5.07 \\
qwen/qwen3-coder-flash & 43.46 & 84.33 & +40.87 \\
qwen/qwen3-coder-plus & 22.99 & 88.94 & +65.95 \\
honeybee-13b & 22.12 & 23.04 & +0.92 \\
qwen/qwen-2.5-coder-32b-instruct & 21.03 & 27.01 & +5.98 \\
honeybee-7b & 19.35 & 23.04 & +3.69 \\
llava-v1.5-7b & 17.97 & 23.96 & +5.99 \\
llamat-2 & 16.59 & 19.82 & +3.23 \\
\bottomrule
\end{tabular}
}
\end{table}

\subsection{Open-Book Mode Observations Using AI-Generated Materials with Expert Review}
Building on the findings from the initial open-book evaluations, we introduced expert review to improve the quality and effectiveness of the supporting materials. Three PhD-level crystallography experts were engaged to refine the AI-generated materials. 

Table~\ref{tab:openbook-expert-results} compares model accuracy in closed-book and open-book modes using expert-reviewed supporting materials across our 217-question crystallography benchmark. The results show substantially improved performance, particularly among LLaVA-based models.

Remarkably, specialized coder models show the most dramatic gains, with qwen3-coder-plus (+65.95\%) and qwen3-coder-flash (+42.25\%) achieving exceptional improvements from very low baselines. Among more general-purpose vision-language models,  the largest accuracy gains are observed in LLaVA-v1.6-34B (+11.52\%) and LLaVA-v1.6-mistral-7B (+11.07\%), highlighting the effectiveness of expert-curated guidance in boosting model reasoning.

\begin{figure}[htbp]
    \centering
    \includegraphics[width=\linewidth]{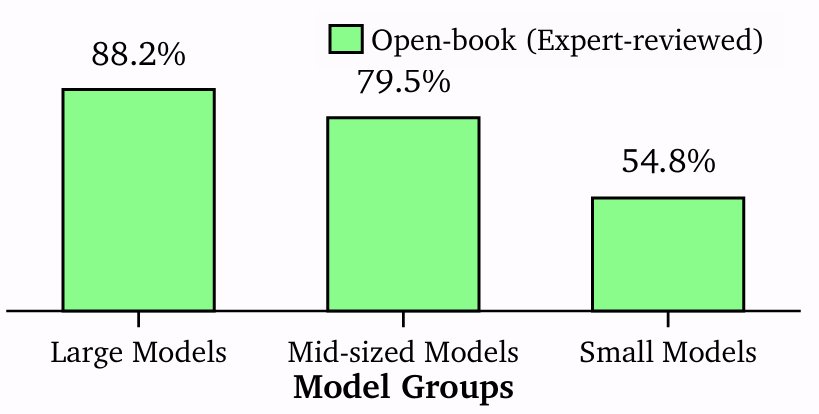}
    \caption{Open-book performance with expert-reviewed supporting materials by model size group. Large Models (>70B parameters or advanced architectures, n=27); Mid-sized Models (>7B-70B parameters, n=22); Small Models ($\leq$7B parameters, n=25). Compared to AI-generated materials (Figure~\ref{fig:openbook_ai_performance}), expert curation provides additional gains for mid-sized ($+1.7\%$) and small models ($+2.5\%$), while large models remain stable, demonstrating that content quality improvements benefit models with reasoning capacity but incomplete domain knowledge.}
    \label{fig:openbook_expert_performance}
\end{figure}

Several other mid-sized models, such as LLaVA-onevision-QWEN2-7B-ov-chat (+10.14\%) and LLaVA-onevision-QWEN2-7B-ov (+8.75\%), also demonstrate notable improvements. Figure~\ref{fig:openbook_expert_performance} illustrates these gains aggregated by model size group, showing that expert curation provides additional improvements over AI-generated materials, particularly for mid-sized ($+1.7\%$) and small models ($+2.5\%$), while large models maintain stable performance.

A notable failure mode identified in our analysis is catastrophic degradation in domain-specialized models, which experience substantial performance drops despite receiving expert-reviewed supporting materials. As summarized in Tables~\ref{tab:openbook-ai-results}--\ref{tab:openbook-expert-results}, the LLaMAT family—pretrained on materials-science literature and domain-focused instruction datasets—shows the most severe declines. LLaMAT-3-chat falls from 57.14\% (closed-book) to 31.80\% (open-book with expert review), a --25.34\% drop, while LLaMAT-2-chat decreases from 50.69\% to 16.13\%, a --34.56\% degradation that renders the model nearly unusable in open-book mode. Only the base llamat-2 model shows a modest improvement (+3.23\%), though its low absolute accuracy indicates fundamental capacity limitations regardless of external support. These results demonstrate that domain specialization alone does not guarantee effective external-knowledge integration.

Similar patterns emerge in other domain-focused families. The HoneyBee models, designed primarily for scientific figure interpretation, show limited transferability to text-only crystallographic reasoning. HoneyBee-13B drops from 22.12\% to 21.20\% (--0.92\%), and HoneyBee-7B from 19.35\% to 17.51\% (--1.84\%), suggesting that pretraining on visual-scientific corpora does not enhance performance on purely textual tasks, and that their small parameter scale (7B--13B) restricts knowledge retention. The dZiner family, consisting of QWEN variants fine-tuned on materials-science corpora, exhibits a similar sensitivity to interference. While dZiner-QWEN-2.5-72B attains strong closed-book accuracy (90.32\%), it drops to 86.18\% (--4.14\%) with expert-reviewed materials. The medium-sized dZiner-QWEN-2.5-coder-32B declines by --6.45\%, and the smaller dZiner-QWEN-2.5-7B by --2.31\%. These negative open-book effects mirror degradation trends observed in other high-capacity models, indicating that domain-specific pretraining raises baseline performance but simultaneously increases vulnerability to information interference.

Across all three families, the underlying mechanism is representational rigidity introduced during domain adaptation. These models internalize crystallographic and materials-science concepts in narrowly defined textual, visual, or instructional formats. When exposed to expert-reviewed passages that express these concepts using more pedagogical, descriptive, or stylistically diverse language, they encounter distributional mismatch between their internal priors and the external context. Even when the supplemental information is factually correct, this mismatch triggers interference rather than assistance, leading to performance degradation. This reveals a fundamental trade-off in domain specialization: while narrow, domain-focused pretraining can strengthen closed-book accuracy, it reduces flexibility in assimilating new or differently framed knowledge sources.

Interestingly, the knowledge-interference failure pattern extends systematically across top-tier models, as explained in Section~\ref{sec:openbook-ai}. As summarized in Table~\ref{tab:openbook-expert-results}, in the transition from the closed-book setting to the open-book mode with expert-reviewed supporting materials, several frontier systems show measurable degradation despite their high baseline accuracy. GPT-4.5-preview declines from 92.63\% to 89.40\% (--3.23 percentage points), while other state-of-the-art models exhibit comparable or greater decreases: openai/gpt-5 (--3.68\%), x-ai/grok-4-fast (--5.78\%), openai/gpt-5-codex (--4.14\%), and O3-mini (--3.69\%). Importantly, all of these systems exceed 90\% closed-book accuracy, confirming that they already possess extensive crystallographic knowledge. The consistency of this degradation across diverse architectures (spanning distinct training paradigms, model families, and organizations) indicates that knowledge interference is an intrinsic characteristic of high-capacity language models rather than a model-specific artifact. These models appear highly sensitive to redundant or overlapping external information: when provided with additional context that reformulates or reiterates knowledge already internalized, they can misallocate attention, resulting in subtle yet systematic declines in performance even when the supplemental material is accurate and expert-curated.

\begin{table}[t]
\centering
\caption{Comparison of model accuracy in closed-book mode vs.\ open-book mode with expert-reviewed supporting materials. $\Delta =$ (Open-book mode) $-$ (Closed-book mode).}

\resizebox{\columnwidth}{!}{%
\begin{tabular}{l c c c}
\toprule
\textbf{Model} & \textbf{Closed-book mode (\%)} & \textbf{Open-book mode (\%)} & $\Delta$ \\
\midrule
openai/gpt-5 & 96.77 & 93.09 & -3.68 \\
x-ai/grok-4-fast & 96.31 & 90.53 & -5.78 \\
openai/gpt-5-codex & 96.31 & 92.17 & -4.14 \\
google/gemini-2.5-pro & 95.39 & 92.63 & -2.76 \\
o3-mini & 93.55 & 89.86 & -3.69 \\
meituan/longcat-flash-chat & 93.09 & 89.86 & -3.23 \\
gpt-4.5-preview & 92.63 & 89.40 & -3.23 \\
anthropic/claude-3.5-sonnet & 91.24 & 89.40 & -1.84 \\
openai/gpt-4o & 90.74 & 87.10 & -3.64 \\
dziner-qwen-2.5-72b & 90.32 & 86.18 & -4.14 \\
qwen/qwen3-next-80b-a3b-instruct & 90.32 & 89.40 & -0.92 \\
perplexity/sonar-pro & 90.32 & 88.02 & -2.30 \\
deepseek/deepseek-v3.1-terminus & 89.86 & 90.78 & +0.92 \\
qwen/qwen-plus & 89.86 & 89.40 & -0.46 \\
qwen/qwen3-next-80b-a3b-thinking & 89.86 & 87.10 & -2.76 \\
deepseek/deepseek-chat & 89.40 & 90.32 & +0.92 \\
qwen/qwen-max & 89.40 & 89.40 & +0.00 \\
o1 & 89.40 & 88.94 & -0.46 \\
anthropic/claude-3-opus & 88.94 & 88.94 & +0.00 \\
openai/o1-mini & 88.02 & 88.02 & +0.00 \\
meta-llama/llama-3.1-405b-instruct & 87.56 & 84.33 & -3.23 \\
qwen/qwen-2.5-72b-instruct & 87.56 & 85.71 & -1.85 \\
mistralai/mistral-large & 86.18 & 86.18 & +0.00 \\
amazon/nova-pro-v1 & 86.11 & 86.64 & +0.53 \\
gpt-4-turbo & 85.25 & 85.71 & +0.46 \\
meta-llama/llama-3-70b-instruct & 84.79 & 82.95 & -1.84 \\
meta-llama/llama-3.1-70b-instruct & 84.79 & 85.19 & +0.40 \\
dziner-qwen-2.5-coder-32b & 83.87 & 77.42 & -6.45 \\
openai/gpt-4-0314 & 83.41 & 82.49 & -0.92 \\
gpt-4-turbo-preview & 83.41 & 84.79 & +1.38 \\
amazon/nova-lite-v1 & 82.03 & 85.71 & +3.68 \\
dziner-qwen-2.5-7b & 81.57 & 79.26 & -2.31 \\
gpt-4 & 81.57 & 82.49 & +0.92 \\
openai/gpt-4o-mini & 81.11 & 81.11 & +0.00 \\
openrouter/auto & 81.11 & 81.57 & +0.46 \\
qwen/qwen-2.5-7b-instruct & 79.72 & 78.34 & -1.38 \\
google/gemma-2-27b-it & 79.26 & 79.72 & +0.46 \\
anthropic/claude-3.5-haiku & 77.57 & 85.92 & +8.35 \\
mistralai/mixtral-8x22b-instruct & 76.81 & 83.89 & +7.08 \\
mistralai/mistral-7b-instruct & 75.35 & 80.65 & +5.30 \\
google/gemma-2-9b-it & 75.12 & 81.57 & +6.45 \\
amazon/nova-micro-v1 & 74.65 & 80.65 & +6.00 \\
mistralai/mistral-small & 74.65 & 83.41 & +8.76 \\
anthropic/claude-3-haiku & 74.04 & 83.33 & +9.29 \\
openai/gpt-3.5-turbo-16k & 72.60 & 75.85 & +3.25 \\
mistralai/mixtral-8x7b-instruct & 72.33 & 77.14 & +4.81 \\
meta-llama/llama-3-8b-instruct & 71.89 & 75.58 & +3.69 \\
mistralai/pixtral-12b & 71.83 & 77.88 & +6.05 \\
openai/gpt-3.5-turbo & 70.67 & 76.33 & +5.66 \\
meta-llama/llama-3.1-8b-instruct & 69.12 & 78.34 & +9.22 \\
llava-v1.6-34b & 66.82 & 78.34 & +11.52 \\
lmms-lab/llava-onevision-qwen2-7b-si & 66.36 & 75.12 & +8.76 \\
lmms-lab/llava-onevision-qwen2-7b-ov-chat & 65.90 & 76.04 & +10.14 \\
lmms-lab/llava-onevision-qwen2-7b-ov & 65.44 & 74.19 & +8.75 \\
meta-llama/llama-3.2-3b-instruct & 64.98 & 65.44 & +0.46 \\
arcee-ai/afm-4.5b & 62.21 & 67.28 & +5.07 \\
mistralai/mistral-7b-instruct-v0.1 & 59.50 & 62.44 & +2.94 \\
perplexity/sonar & 59.24 & 67.14 & +7.90 \\
undi95/remm-slerp-l2-13b & 57.08 & 65.12 & +8.04 \\
llamat-3-chat & 57.14 & 31.80 & -25.34 \\
alibaba/tongyi-deepresearch-30b-a3b & 55.76 & 86.64 & +30.88 \\
gryphe/mythomax-l2-13b & 53.77 & 66.51 & +12.74 \\
llava-v1.6-mistral-7b & 52.99 & 64.06 & +11.07 \\
llamat-2-chat & 50.69 & 16.13 & -34.56 \\
lmms-lab/llava-onevision-qwen2-0.5b-ov & 47.47 & 54.38 & +6.91 \\
llava-v1.5-13b & 46.54 & 48.39 & +1.85 \\
lmms-lab/llava-onevision-qwen2-0.5b-si & 46.08 & 53.91 & +7.83 \\
qwen/qwen3-coder-flash & 43.46 & 85.71 & +42.25 \\
qwen/qwen3-coder-plus & 22.99 & 88.94 & +65.95 \\
honeybee-13b & 22.12 & 21.20 & -0.92 \\
qwen/qwen-2.5-coder-32b-instruct & 21.03 & 30.77 & +9.74 \\
honeybee-7b & 19.35 & 17.51 & -1.84 \\
llava-v1.5-7b & 17.97 & 28.57 & +10.60 \\
llamat-2 & 16.59 & 19.82 & +3.23 \\
\bottomrule
\end{tabular}
}
\label{tab:openbook-expert-results}
\end{table}
\subsection{Ablation Study: Content Quality vs. Information Quantity}
\label{sec:quality_quantity}

To rigorously test whether performance gains stem from information quantity or expert curation quality, we conducted a comprehensive token-matched comparison between AI-generated and expert-reviewed supporting materials. \textit{Tokens} are the fundamental units of text that language models process—roughly corresponding to words or sub-word pieces (e.g., ``crystallography'' might be split into ``crystal'' + ``lography''). By controlling token count, we ensure that both material types contain equivalent amounts of text, allowing us to isolate content quality effects from simple volume differences. This ablation directly addresses the question: do expert improvements reflect better content or simply more text?

\subsubsection{Token-Matched Experimental Design}
To isolate the effects of content quality from those of information quantity, we implemented a token-matched ablation study in which every AI-generated supporting passage was paired with its expert-reviewed counterpart of nearly identical length. This design ensured that both sets of materials provided the same textual volume, allowing any observed performance differences to be directly attributed to qualitative factors such as accuracy, relevance, and pedagogical clarity.

As summarized in Table~\ref{tab:token_stats}, the mean token count was $566.0 \pm 79.2$ for AI-generated materials and $568.9 \pm 82.7$ for expert-reviewed materials, representing a negligible difference of only 2.9 tokens (0.51\% change). Moreover, the high correlation ($r = 0.948$) between AI- and expert-reviewed helper lengths across all 217 questions, along with their nearly identical median and interquartile ranges ($558.0$ [$508.0$--$605.0$] vs. $558.0$ [$513.0$--$610.0$]), confirms that the two distributions are tightly matched.

This strict token control eliminates text volume as a potential confounding factor and establishes a high-fidelity basis for quality-driven comparison. In other words, any measurable performance improvements can be confidently attributed to the qualitative enhancements introduced through expert curation, rather than to variations in the amount of information provided. This methodological precision is critical for disentangling the true impact of expert refinement from superficial gains that might otherwise arise from longer or more verbose passages.

\begin{table}[H]
\centering
\caption{Token Count Statistics for Questions and Helper Texts. Expert-reviewed materials maintain virtually identical length to AI-generated materials (mean difference: 2.9 tokens, 0.51\%), enabling quality-controlled comparison.}
\label{tab:token_stats}

\resizebox{\columnwidth}{!}{%
\begin{tabular}{lrrr}
\toprule
\textbf{Metric} & \textbf{Questions} & \textbf{AI-Generated} & \textbf{Expert-Reviewed} \\
\midrule
Mean $\pm$ SD & $50.1 \pm 15.2$ & $566.0 \pm 79.2$ & $568.9 \pm 82.7$ \\
Median [IQR] & $49.0$ [38.0--60.0] & $558.0$ [508.0--605.0] & $558.0$ [513.0--610.0] \\
Range & [21--98] & [378--919] & [378--919] \\
Total & 10,882 & 122,825 & 123,457 \\
\midrule
\textbf{Difference} & — & — & \textbf{+2.9 tokens (+0.51\%)} \\
\textbf{Correlation} & — & — & \textbf{$r=0.948$} \\
\bottomrule
\end{tabular}%
}
\end{table}

\subsubsection{LLM-Based Helper Quality Scoring}
\label{sec:llm-quality-ratings}

To quantitatively validate the qualitative advantages of expert-reviewed helpers, we scored all 217 AI-generated vs.\ human-reviewed helper pairs using an external rater (GPT-4o-mini via OpenRouter). The rater assigned 0--10 scores on four dimensions aligned with our quality rubric: accuracy, clarity, completeness, and reliability. Human-reviewed passages outperform AI-generated passages by +1.38 to +1.87 points per metric and by +1.72 points on the overall average (Table~\ref{tab:helper-quality-ratings}), confirming that expert edits measurably increase helper quality beyond text length controls.

\begin{table}[t]
\centering
\caption{LLM-based quality ratings (0--10) for 217 helper pairs. Scores are means across questions.}
\label{tab:helper-quality-ratings}
\begin{tabular}{lccc}
\toprule
\textbf{Metric} & \textbf{AI helper} & \textbf{Human helper} & $\boldsymbol{\Delta}$ \\
\midrule
Accuracy      & 7.43 & 9.25 & +1.82 \\
Clarity       & 7.53 & 8.91 & +1.38 \\
Completeness  & 7.40 & 9.20 & +1.81 \\
Reliability   & 7.30 & 9.17 & +1.87 \\
\midrule
Overall mean  & 7.41 & 9.13 & +1.72 \\
\bottomrule
\end{tabular}
\end{table}

\subsubsection{Differential Performance Under Token-Matched Conditions}

Table~\ref{tab:comparative-results} presents a direct comparison of performance gains ($\Delta$) between token-matched AI-generated and expert-reviewed supporting materials across all 74 evaluated models. Despite identical token counts, substantial performance differences emerge, revealing the effect of content quality independent of text volume.

Most models show higher accuracy with expert-reviewed materials, confirming that qualitative improvements—such as enhanced conceptual clarity, precise terminology, and improved contextual alignment—produce measurable performance gains even under identical token counts. The strongest relative improvements occur in small models (<7B parameters), which benefit most from the additional clarity and precision introduced by expert curation. For instance, LLaVA-v1.6-mistral-7B improves from +5.54\% to +11.07\% (+5.53\% differential), and mistralai/mistral-7b-instruct-v0.1 exhibits the largest quality-driven transformation (+8.41\%), shifting from performance degradation (--5.47\%) with AI-generated text to a measurable gain (+2.94\%) with expert-reviewed material, under identical token budgets.

Mid-sized models (10–70B) also show clear but smaller improvements, exemplified by LLaVA-v1.6-34B (+5.99\% to +11.52\%, +5.53\% differential). These models possess enough reasoning capacity to leverage higher-quality information but already contain partial crystallographic knowledge, so their absolute gains are moderate.

Conversely, large models (>70B), such as GPT-4.5-preview, O3-mini, and O1, exhibit minimal or slightly negative differences, reflecting a saturation of internal knowledge representations. When most relevant crystallographic concepts are already encoded within the model parameters, additional context contributes little new information and can occasionally disrupt attention alignment or introduce representational interference.

Overall, these results confirm that content quality, not text length, governs open-book performance. Expert review yields the greatest relative benefits for small and mid-sized models, while large models approach a ceiling where additional context produces diminishing or even adverse effects.

Figure~\ref{fig:performance_gains_comparison} summarizes the overall impact of supporting material quality across model size categories. Small models (<7B parameters) achieve the largest relative improvement from expert-reviewed materials, increasing their accuracy by +8.52 percentage points compared to +6.18\% with AI-generated content. Mid-sized models (7B–70B) also benefit noticeably, showing a +4.44\% gain versus +2.71\% with AI-generated materials. In contrast, large models (>70B) display minimal or slightly negative changes with either material type, indicating that their extensive internal knowledge already limits the marginal benefit of added context.

This trend can be explained by differences in knowledge representation and context utilization capacity across model scales. Smaller models lack sufficient internal crystallographic knowledge, so expert-reviewed passages provide missing conceptual structure and precise terminology that directly enhance reasoning accuracy. Mid-sized models already encode partial domain knowledge but still rely on external information to fill gaps, making them responsive to high-quality supporting text. In contrast, very large models possess extensive parametric knowledge, and adding redundant or overlapping context can introduce interference effects, where new information competes with their established internal representations. As a result, they gain little additional benefit and may even experience slight degradation when the external input does not perfectly align with their internal reasoning pathways.

Overall, these findings indicate that while expert curation improves performance across all model scales, the relative gains are most pronounced for small and mid-sized models that can still meaningfully integrate external guidance, whereas large models operate near their knowledge saturation limit. 

\paragraph{Practical significance of Fig.~\ref{fig:performance_gains_comparison}.} Improving average accuracy from 72.5\% to 79.5\% (Fig.~\ref{fig:performance_gains_comparison}) is a 7.0 percentage-point gain that lowers the error rate from 27.5\% to 20.5\%—about a 25\% reduction in wrong answers, or roughly 15 additional correct answers on our 217-question benchmark. Moving from ~3 wrong answers in 10 to ~2 can be noticeable to users because it reduces re-checking and re-asking. However, 79.5\% still means ~1 in 5 answers is incorrect, so high-stakes crystallography (e.g., indexing, space-group interpretation, experimental planning) should remain human-verified. The main benefit is cost-effective assistance: mid-sized models plus expert-reviewed hints approach large-model accuracy at lower compute cost.

\subsubsection{Quantity-Driven Effects: Evidence from Token Budget Constraints}

To further disentangle quality from quantity effects, Section~\ref{sec:token-constraints} presents complementary evidence that information quantity alone can degrade performance. Our token budget analysis (Figure~\ref{fig:token_budget_effects}) reveals that llamat-3-chat's open-book accuracy dropped from 36.87\% at 256 tokens to 26.73\% at 2048+ tokens—a 10 percentage point degradation—when provided with longer supporting materials. This inverse relationship demonstrates severe token starvation: the model exhausts its computational budget on input processing, leaving insufficient capacity for answer generation. The supporting materials, rather than aiding reasoning, become a computational burden that crowds out response formulation.

This finding establishes a critical control, if performance gains stemmed primarily from text volume rather than content quality, we would not observe degradation with increased token budgets in resource-constrained models. The token starvation phenomenon proves that ``more text'' does not guarantee better outcomes and, in fact, can be counterproductive when models lack the capacity to efficiently process additional context.

\subsubsection{Interpretation: Quality, Not Quantity, Drives Improvements}

The combination of token-matched comparisons (identical length, differential outcomes) and token budget analysis (increased length, degraded outcomes) provides convergent evidence that content quality—specifically, accuracy, relevance, and pedagogical clarity—drives the observed performance improvements, not mere text volume. Expert-reviewed materials successfully addressed the limitations identified in AI-generated content: technical inaccuracies, insufficient depth, tangential information, and suboptimal pedagogical framing. These qualitative enhancements enable mid-capacity models to better integrate external knowledge, bridging the gap between their processing capability and incomplete domain expertise.

Critically, the token-matched design ensures that expert improvements cannot be attributed to providing "more information" in terms of token count. Instead, expert review enhances information density, precision, and alignment with question-specific reasoning needs—dimensions of quality orthogonal to text length. This ablation establishes that the OPENXRD framework's performance gains reflect genuine knowledge enhancement rather than computational artifacts of longer context windows.

\begin{table}[t]
\centering
\caption{Direct comparison of performance gains between AI-generated supporting materials with and without expert review.}
\label{tab:comparative-results}
\resizebox{\columnwidth}{!}{%
\begin{tabular}{l c c c}
\toprule
\textbf{Model} & \textbf{AI-Generated $\Delta$ (\%)} & \textbf{Expert-Reviewed $\Delta$ (\%)} & \textbf{Difference} \\
\midrule
openai/gpt-5 & -1.84 & -3.68 & -1.84 \\
x-ai/grok-4-fast & -0.92 & -5.78 & -4.86 \\
openai/gpt-5-codex & -1.84 & -4.14 & -2.30 \\
google/gemini-2.5-pro & -6.91 & -2.76 & +4.15 \\
o3-mini & -4.15 & -3.69 & +0.46 \\
meituan/longcat-flash-chat & -2.77 & -3.23 & -0.46 \\
gpt-4.5-preview & -2.31 & -3.23 & -0.92 \\
anthropic/claude-3.5-sonnet & -1.84 & -1.84 & +0.00 \\
openai/gpt-4o & -3.18 & -3.64 & -0.46 \\
dziner-qwen-2.5-72b & -3.22 & -4.14 & -0.92 \\
qwen/qwen3-next-80b-a3b-instruct & -0.92 & -0.92 & +0.00 \\
perplexity/sonar-pro & +0.00 & -2.30 & -2.30 \\
deepseek/deepseek-v3.1-terminus & +0.92 & +0.92 & +0.00 \\
qwen/qwen-plus & +0.92 & -0.46 & -1.38 \\
qwen/qwen3-next-80b-a3b-thinking & -2.76 & -2.76 & +0.00 \\
deepseek/deepseek-chat & +1.84 & +0.92 & -0.92 \\
qwen/qwen-max & -1.38 & +0.00 & +1.38 \\
o1 & -1.38 & -0.46 & +0.92 \\
anthropic/claude-3-opus & -0.92 & +0.00 & +0.92 \\
openai/o1-mini & +0.46 & +0.00 & -0.46 \\
meta-llama/llama-3.1-405b-instruct & -4.61 & -3.23 & +1.38 \\
qwen/qwen-2.5-72b-instruct & -1.38 & -1.85 & -0.47 \\
mistralai/mistral-large & -2.31 & +0.00 & +2.31 \\
amazon/nova-pro-v1 & +0.53 & +0.53 & +0.00 \\
gpt-4-turbo & +0.46 & +0.46 & +0.00 \\
meta-llama/llama-3-70b-instruct & -4.14 & -1.84 & +2.30 \\
meta-llama/llama-3.1-70b-instruct & -1.38 & +0.40 & +1.78 \\
dziner-qwen-2.5-coder-32b & +0.92 & -6.45 & -7.37 \\
openai/gpt-4-0314 & -5.99 & -0.92 & +5.07 \\
gpt-4-turbo-preview & -0.92 & +1.38 & +2.30 \\
amazon/nova-lite-v1 & +2.76 & +3.68 & +0.92 \\
dziner-qwen-2.5-7b & -2.31 & -2.31 & +0.00 \\
gpt-4 & +0.46 & +0.92 & +0.46 \\
openai/gpt-4o-mini & +0.92 & +0.00 & -0.92 \\
openrouter/auto & +0.46 & +0.46 & +0.00 \\
qwen/qwen-2.5-7b-instruct & -4.60 & -1.38 & +3.22 \\
google/gemma-2-27b-it & -0.46 & +0.46 & +0.92 \\
anthropic/claude-3.5-haiku & +10.45 & +8.35 & -2.10 \\
mistralai/mixtral-8x22b-instruct & +3.28 & +7.08 & +3.80 \\
mistralai/mistral-7b-instruct & +2.07 & +5.30 & +3.23 \\
google/gemma-2-9b-it & +2.30 & +6.45 & +4.15 \\
amazon/nova-micro-v1 & +4.61 & +6.00 & +1.39 \\
mistralai/mistral-small & +4.61 & +8.76 & +4.15 \\
anthropic/claude-3-haiku & +8.45 & +9.29 & +0.84 \\
openai/gpt-3.5-turbo-16k & -1.59 & +3.25 & +4.84 \\
mistralai/mixtral-8x7b-instruct & +3.63 & +4.81 & +1.18 \\
meta-llama/llama-3-8b-instruct & +1.38 & +3.69 & +2.31 \\
mistralai/pixtral-12b & +3.75 & +6.05 & +2.30 \\
openai/gpt-3.5-turbo & +0.83 & +5.66 & +4.83 \\
meta-llama/llama-3.1-8b-instruct & +6.00 & +9.22 & +3.22 \\
llava-v1.6-34b & +5.99 & +11.52 & +5.53 \\
lmms-lab/llava-onevision-qwen2-7b-si & +5.53 & +8.76 & +3.23 \\
lmms-lab/llava-onevision-qwen2-7b-ov-chat & +6.45 & +10.14 & +3.69 \\
lmms-lab/llava-onevision-qwen2-7b-ov & +5.99 & +8.75 & +2.76 \\
meta-llama/llama-3.2-3b-instruct & -2.31 & +0.46 & +2.77 \\
arcee-ai/afm-4.5b & +2.31 & +5.07 & +2.76 \\
mistralai/mistral-7b-instruct-v0.1 & -5.47 & +2.94 & +8.41 \\
perplexity/sonar & +9.60 & +7.90 & -1.70 \\
undi95/remm-slerp-l2-13b & +4.42 & +8.04 & +3.62 \\
llamat-3-chat & -26.73 & -25.34 & +1.39 \\
alibaba/tongyi-deepresearch-30b-a3b & +24.42 & +30.88 & +6.46 \\
gryphe/mythomax-l2-13b & +7.91 & +12.74 & +4.83 \\
llava-v1.6-mistral-7b & +5.54 & +11.07 & +5.53 \\
llamat-2-chat & -29.56 & -34.56 & -5.00 \\
lmms-lab/llava-onevision-qwen2-0.5b-ov & +3.68 & +6.91 & +3.23 \\
llava-v1.5-13b & -2.30 & +1.85 & +4.15 \\
lmms-lab/llava-onevision-qwen2-0.5b-si & +5.07 & +7.83 & +2.76 \\
qwen/qwen3-coder-flash & +40.87 & +42.25 & +1.38 \\
qwen/qwen3-coder-plus & +65.95 & +65.95 & +0.00 \\
honeybee-13b & +0.92 & -0.92 & -1.84 \\
qwen/qwen-2.5-coder-32b-instruct & +5.98 & +9.74 & +3.76 \\
honeybee-7b & +3.69 & -1.84 & -5.53 \\
llava-v1.5-7b & +5.99 & +10.60 & +4.61 \\
llamat-2 & +3.23 & +3.23 & +0.00 \\
\bottomrule
\end{tabular}%
}
\label{tab:comparative-results}
\end{table}

\begin{figure}[htbp]
\centering
\begin{tikzpicture}
\begin{axis}[
    width=\columnwidth,
    height=4.5cm,
    symbolic x coords={Large Models, Mid-sized Models, Small Models},
    xtick=data,
    ybar=2pt,
    bar width=28pt,
    ymin=-10.0, ymax=12,
    enlarge x limits=0.15,
    axis x line* = bottom,
    axis y line* = left,
    axis line style={black, line width=0.7pt},
    tick style={line width=0.7pt, black},
    nodes near coords={\pgfmathprintnumber[fixed, precision=2]\pgfplotspointmeta\%},
    every node near coord/.append style={font=\normalsize, black, anchor=south, yshift=2pt},
    xticklabel style={font=\small},
    xlabel style={font=\small, font=\bfseries, yshift=5pt},
    ytick=\empty,
    y axis line style={draw=none},
    grid=none,
    legend style={
        at={(0.02,1.15)},
        anchor=south west,
        legend columns=-1,
        draw=none,
        font=\small,
        cells={anchor=west},
        column sep=8pt
    },
    clip=false,
    every axis plot/.append style={
        legend image code/.code={
            \draw[#1, draw=black] (0cm,-0.1cm) rectangle (0.3cm,0.15cm);
        }
    }
]
\addplot[
    fill=orange!35,
    draw=black,
    line width=0.8pt
] coordinates {
    (Large Models, -1.28)
    (Mid-sized Models, 2.71)
    (Small Models, 6.18)
};
\addlegendentry{AI-Generated Materials}
\addplot[
    fill=teal!35,
    draw=black,
    line width=0.8pt
] coordinates {
    (Large Models, -1.77)
    (Mid-sized Models, 4.44)
    (Small Models, 8.52)
};
\addlegendentry{Expert-Reviewed Materials}
\end{axis}
\end{tikzpicture}
\caption{Average performance gains from AI-generated vs. expert-reviewed supporting materials across model size categories. Large Models ($>$70B parameters or advanced architectures like GPT-4, GPT-5, O1, O3-mini, n=27); Mid-sized Models ($>$7B-70B parameters including LLaVA-34B, QWEN2-7B, Mistral variants, n=27); Small Models ($\leq$7B parameters including QWEN2-0.5B, LLaMA-3.2-3B, LLaVA-7B, LLaMAT variants, Honeybee-7B, n=20). Note that large models show performance degradation with supporting materials, while mid-sized and small models benefit substantially from expert-reviewed materials. Overall, these results show that as models become larger and contain more built-in knowledge, the additional benefit from external information gradually diminishes.}
\label{fig:performance_gains_comparison}
\end{figure}

\subsection{Detailed Subtask-Level Performance with Expert-Reviewed Materials}
\label{sec:subtask-analysis}
To gain deeper insights, we examined \textit{subtask-level} performance with expert-reviewed materials. Table~\ref{tab:subtask-improvements} shows how LLaVA-onevision-QWEN2-7B-ov behaves on key XRD subtasks. The rationale for selecting this model is its substantial enhancement from expert-reviewed materials, which offers valuable insights into the limits of performance improvement. 
For several subtasks such as Crystal Structure, Laue Patterns, Metal Structures, and Powder Diffraction, expert-reviewed materials transformed performance from complete failure (0\%) to perfect accuracy (100\%). Expert materials particularly improved performance on structural tasks, enhancing Structure Factors by 25\% and Coordination Numbers by 40\%. However, complex mathematical derivations like Bragg's Law and Calculation Methods remained challenging even with expert materials, suggesting that some concepts require more than textual explanation.
An illustrative example is the question "\emph{What is the structure factor $F$ for a base-centered unit cell when $h$ and $k$ are mixed (one even, one odd)?}" Correct answer is being $F=f$, while a model consistently selecting $F=2f$, reveals that it has not internalized how lattice symmetry produces systematic absences,  an error that echoes its 0\% score on Bragg's Law questions.
Additionally, for a few subtasks where the model had strong initial performance, such as Complex Mathematics and Wave Scattering, supporting materials appeared to interfere with the model's existing knowledge.

Mathematically intensive subtasks including Bragg's Law derivations, structure factor calculations, and reflection condition analysis showed universal 0\% improvement with supporting materials across all 74 models. For instance, when asked about structure factors for base-centered lattices with mixed h,k indices, models consistently failed to execute the symbolic computation $F = f[1 + e^{i\pi(h+k)}]$ despite correct textual explanations, revealing that current LLMs cannot perform the formal algebraic operations required for crystallographic problem-solving.

\begin{table}[ht]
\centering
\caption{Selected subtask-level accuracy (\%) for LLaVA-onevision-QWEN2-7B-ov, comparing closed-book mode vs.\ open-book mode with expert-reviewed materials. Some subtasks have fewer questions, so a single item shifts accuracy considerably.}
\label{tab:subtask-improvements}
\vspace{0.5em}

\resizebox{0.85\columnwidth}{!}{%
\begin{tabular}{lccc}
\toprule
\textbf{Subtask} & \textbf{Closed-book mode} & \textbf{Open-book mode} & \textbf{Improvement} \\
\midrule
\textit{Large Gains} & & & \\
\quad Atomic Spacing           & 66.7 & 100.0 & +33.3 \\
\quad Coordination Numbers     & 20.0 & 60.0  & +40.0 \\
\quad Crystal Structure        &  0.0 & 100.0 & +100.0 \\
\quad Laue Patterns            &  0.0 & 100.0 & +100.0 \\
\quad Metal Structures         &  0.0 & 100.0 & +100.0 \\
\quad Powder Diffraction       &  0.0 & 100.0 & +100.0 \\
\quad Structure Factors        & 60.0 & 85.0  & +25.0 \\
\midrule
\textit{Little / No Improvement} & & & \\
\quad Bragg's Law                &  0.0 &  0.0  & +0.0 \\
\quad Calculation Methods      &  0.0 &  0.0  & +0.0 \\
\quad Diffraction Limitations  &  0.0 &  0.0  & +0.0 \\
\midrule
\textit{Performance Drops} & & & \\
\quad Complex Mathematics       & 100.0 & 50.0 & -50.0 \\
\quad Wave Scattering          & 100.0 &  0.0 & -100.0 \\
\bottomrule
\end{tabular}
}
\end{table}

\subsection{Impact of Token Budget Constraints on Open-Book Performance}
\label{sec:token-constraints}

While the previous analyses examined model performance under standard inference conditions, we now investigate how computational constraints—specifically, output token budget limitations—affect models' ability to leverage external knowledge. Token budget restrictions are particularly relevant in deployment scenarios where computational resources or API costs impose practical limits on generation length. Understanding how these constraints interact with open-book augmentation is crucial for real-world applications.

We systematically evaluated two models across varying maximum token budgets (256 to 4096 tokens): llamat-3-chat, a 7B-parameter specialized model, and O3-mini, a reasoning-optimized model designed for scientific tasks. Figure~\ref{fig:token_budget_effects} presents the accuracy trends as token budgets increase.

\begin{figure*}[t]
\centering
\resizebox{0.95\textwidth}{!}{%
\begin{tikzpicture}
\begin{axis}[
    width=0.48\textwidth,
    height=6cm,
    xlabel={Max Tokens},
    ylabel={Accuracy (\%)},
    xmin=0, xmax=4500,
    ymin=20, ymax=100,
    xtick={256,512,768,1024,1536,2048,2560,3072,3584,4096},
    xticklabel style={rotate=45, anchor=east, font=\footnotesize},
    legend style={at={(0.5,-0.35)},anchor=north,legend columns=2,font=\footnotesize},
    grid=major,
    grid style={dashed,gray!30},
    title={\textbf{(a) llamat-3-chat}},
    title style={font=\small}
]
\addplot[color=blue,mark=square*,thick,mark size=2.5pt] coordinates {
    (256,55.30) (512,55.30) (768,57.60) (1024,58.53) 
    (1536,55.76) (2048,56.68) (2560,57.14) (3072,57.14) 
    (3584,57.14) (4096,56.22)
};
\addlegendentry{Closed-book}

\addplot[color=red,mark=*,thick,mark size=2.5pt] coordinates {
    (256,36.87) (512,29.49) (768,31.34) (1024,30.41) 
    (1536,30.41) (2048,26.73) (2560,26.73) (3072,26.73) 
    (3584,26.73) (4096,26.73)
};
\addlegendentry{Open-book}
\end{axis}
\end{tikzpicture}
\hfill
\begin{tikzpicture}
\begin{axis}[
    width=0.48\textwidth,
    height=6cm,
    xlabel={Max Tokens},
    ylabel={Accuracy (\%)},
    xmin=0, xmax=4500,
    ymin=70, ymax=100,
    xtick={256,512,768,1024,1536,2048,2560,3072,3584,4096},
    xticklabel style={rotate=45, anchor=east, font=\footnotesize},
    legend style={at={(0.5,-0.35)},anchor=north,legend columns=2,font=\footnotesize},
    grid=major,
    grid style={dashed,gray!30},
    title={\textbf{(b) O3-mini}},
    title style={font=\small}
]
\addplot[color=blue,mark=square*,thick,mark size=2.5pt] coordinates {
    (256,77.42) (512,85.71) (768,89.40) (1024,91.24) 
    (1536,94.01) (2048,94.93) (2560,95.39) (3072,93.55) 
    (3584,94.01) (4096,93.55)
};
\addlegendentry{Closed-book}

\addplot[color=red,mark=*,thick,mark size=2.5pt] coordinates {
    (256,80.65) (512,88.94) (768,88.94) (1024,88.94) 
    (1536,90.32) (2048,90.32) (2560,90.32) (3072,90.78) 
    (3584,89.86) (4096,90.32)
};
\addlegendentry{Open-book}
\end{axis}
\end{tikzpicture}
}
\caption{Impact of token budget constraints on closed-book and open-book performance for (a) llamat-3-chat and (b) O3-mini. The x-axis shows maximum output tokens allowed, while the y-axis shows accuracy on the 217-question benchmark. Note the different y-axis scales: llamat-3-chat ranges 20-100\% while O3-mini ranges 70-100\% due to their different baseline capabilities.}
\label{fig:token_budget_effects}

\end{figure*}

The results reveal markedly different patterns between the two models. For llamat-3-chat (Figure~\ref{fig:token_budget_effects}a), closed-book performance remains relatively stable across token budgets (55-58\%), showing minimal sensitivity to generation length constraints. However, open-book performance exhibits a counterintuitive decline: accuracy drops from 36.87\% at 256 tokens to 26.73\% at 2048+ tokens—a degradation of approximately 10 percentage points. This inverse relationship suggests severe token starvation: as the model attempts to process longer supporting materials, it exhausts its token budget on input comprehension and internal reasoning, leaving insufficient capacity for coherent answer generation. The supporting materials, rather than aiding reasoning, become a computational burden that crowds out the model's ability to formulate responses.

In contrast, O3-mini (Figure~\ref{fig:token_budget_effects}b) demonstrates more favorable scaling properties. Closed-book accuracy improves substantially from 77.42\% at 256 tokens to a peak of 95.39\% at 2560 tokens, indicating the model benefits from additional reasoning space for complex crystallographic problems. Open-book performance similarly improves from 80.65\% to approximately 90\%, stabilizing around 1024 tokens. The gap between closed-book and open-book performance narrows at higher token budgets, suggesting that O3-mini can more effectively integrate external knowledge when given adequate computational headroom. However, even O3-mini shows slight performance degradation beyond 2560 tokens in closed-book mode, hinting at potential overfitting or unnecessary elaboration when unconstrained.

These findings have important implications for deployment strategies. First, token budget selection requires careful calibration: smaller models like llamat-3-chat exhibit pathological behavior under open-book conditions when token budgets exceed their processing capacity, while more capable models like O3-mini require minimum token budgets (~1024 tokens) to effectively leverage external materials. Second, the phenomenon of "open-book degradation" in resource-constrained models suggests that naïvely providing supporting materials without ensuring sufficient token allocation for answer synthesis can be counterproductive. Third, the results underscore fundamental architectural differences: reasoning-optimized models (O3-mini) appear better equipped to manage the token allocation trade-off between comprehending external context and generating accurate responses.

From a practical standpoint, these observations suggest that effective open-book augmentation requires not only high-quality supporting materials but also appropriate computational provisioning. For deployment scenarios with strict token limits, practitioners should either: (1) use models specifically designed for reasoning tasks that can efficiently allocate tokens between context processing and response generation, (2) employ adaptive token budgets that scale with input complexity, or (3) pre-process supporting materials to reduce their token footprint while preserving essential information. Future work should investigate token allocation strategies that dynamically balance between context comprehension and response generation, potentially through attention mechanism modifications or explicit token budgeting protocols.

\section{Discussion}\label{sec:discussion}
Our results show that using domain-specific prompts, or open-book mode, significantly enhances question-answering accuracy in crystallography. This is especially true for smaller or more general models, where the additional context helps fill knowledge gaps. However, the quality and relevance of the supporting textual material are crucial. In some instances, the material generated by GPT-4.5 did not align well with the question's needs, causing confusion or contradictory information. This underscores the importance of relevance filtering or validation of supporting material before it is used.

While OPENXRD focuses on inference-time context assimilation, it is explicitly designed to complement parameter-efficient fine-tuning (PEFT) approaches. PEFT methods can encode crystallographic knowledge directly into model weights through techniques like LoRA~\cite{hu2021lora}, whereas OPENXRD diagnoses whether models—adapted or not—can effectively utilize answer-guiding context that does not directly reveal solutions. A comprehensive evaluation strategy might compare: (i) closed-book performance after PEFT on external XRD corpora, (ii) open-book performance after PEFT using our oracle passages, and (iii) our baseline open-book performance without PEFT. This three-way comparison decomposes gains attributable to parametric adaptation versus gains from inference-time guidance, a separation difficult to obtain in end-to-end training studies. For instance, a PEFT-adapted model that shows minimal open-book improvement might indicate successful knowledge internalization, while a model showing large open-book gains suggests incomplete adaptation where external guidance remains valuable. OPENXRD standardizes this diagnostic capability across architectures and scales.

Advanced mathematical reasoning remains a challenge. Our analysis indicates that open-book mode rarely resolves complex mathematical problems, such as structure factor calculations or multi-step interference proofs. Future enhancements could include integrating symbolic math modules or domain-specific solvers to better handle these tasks, rather than relying solely on textual references.

The complexity of visual data in crystallography, such as tabulated results, diffraction patterns, and annotated structural diagrams, presents additional challenges. Accurate extraction of this information requires improvements in optical character recognition (OCR), figure parsing, and multi-modal alignment. Developing tailored visual backbones or domain-tuned OCR engines will be essential for applying our methods to real-world scenarios, like analyzing lab notebooks or older textbook scans.

Text-based supporting materials prove insufficient for mathematically intensive subtasks requiring symbolic manipulation. Consider a diagnostic case from our benchmark: ``For a base-centered unit cell, what is the structure factor F when h and k are mixed (one even, one odd)?'' Despite expert-reviewed materials correctly explaining that atoms at (0,0,0) and (1/2,1/2,0) produce destructive interference when h and k have mixed parity (yielding F=0), multiple frontier models incorrectly answered F=2f. Analysis reveals they cannot maintain intermediate symbolic states during phase calculations $\phi = 2\pi(hx + ky + lz)$ or correctly apply $F = f[1 + e^{i\pi(h+k)}] = 0$ when $(h+k)$ is odd. This failure extends to Bragg's Law derivations, structure factor algebra, and powder diffraction indexing—all showing 0\% improvement across our 74-model evaluation regardless of size or architecture.

To overcome this architectural limitation, future work should integrate LLMs with symbolic computation engines such as SymPy~\cite{meurer2017sympy} or Wolfram Alpha~\cite{wolfram2009wolframalpha} for exact algebraic manipulations, crystallographic knowledge graphs encoding systematic absence rules and space group constraints as structured logical representations rather than text, and domain-specific software including GSAS-II~\cite{toby2013gsasii} for powder diffraction refinement and Mercury~\cite{macrae2020mercury} for structure visualization. Such hybrid architectures would enable the LLM to parse problems linguistically, delegate mathematical operations to specialized modules, and synthesize results—combining linguistic flexibility with computational rigor. This approach extends beyond crystallography to other physics-heavy domains requiring both conceptual reasoning and mathematical precision.

In parallel, a notable tension exists between generalist and specialist models. While larger LLMs like GPT-4 and GPT-4.5 have broad knowledge, they often miss the nuances of specific domains that specialized models capture more effectively. A hybrid approach, where a specialist model generates domain-specific references for a generalist model, could offer a powerful solution. Additionally, ethical and copyright considerations must be managed carefully, especially as we look to expand these systems to other technical fields with extensive proprietary literature.

Our research identifies two primary methods for enhancing model performance with domain expertise: directly embedding knowledge into large models, such as frontier models like GPT-5, GPT-4.5, and O3-mini, and providing models with access to expert-reviewed references during inference. This latter strategy proves especially useful for mid-capacity models, enabling them to nearly match the performance of their larger counterparts by leveraging a ``knowledge bridge.''

For example, a medium-sized model like LLaVA-v1.6-34B, when paired with expert materials, achieves 78.34\% accuracy, approaching the performance of much larger models like Llama-3.1-405B (84.33\%) and even surpassing some frontier models in cost-effectiveness but with far fewer resources required. The integration of expert-reviewed domain knowledge is not only beneficial in textual analysis but also in multi-modal contexts such as medical imaging or geospatial analysis, where expert-reviewed context can similarly boost specialized question-answering performance and unlock new application areas.

However, the addition of external information is not always beneficial. For instance, multiple frontier models show performance degradation with expert-reviewed supporting materials: GPT-4.5-preview (-3.23\%), GPT-5 (-3.68\%), Grok-4-fast (-5.78\%), Claude-3.5-Sonnet (-1.84\%), and dziner-qwen-2.5-72b (-4.14\%), possibly due to conflicts with their pre-trained internal knowledge. This suggests that more information is not necessarily better, particularly for top-tier models that have been extensively trained, as the interference from external context can disrupt their already-comprehensive internal reasoning processes.

Our results show that both small and mid-sized models benefit from external knowledge, with small models exhibiting the largest relative gains, whereas large models experience interference. This reveals a fundamental insight: external knowledge augmentation is most effective within a specific capability range where models have sufficient processing power but incomplete domain coverage. This finding has important implications for cost-effective deployment strategies in specialized scientific domains.

Future research should focus on automated methods to determine when expert review is most beneficial, multi-stage approaches for targeted expert review, and the potential for fine-tuning models to better utilize external references, as well as mechanisms to detect when models have sufficient internal knowledge that external context becomes counterproductive.

Beyond incremental improvements, addressing the universal mathematical reasoning failures documented in our benchmark requires architectural innovation. Hybrid systems coupling LLMs with symbolic algebra engines, structured crystallographic knowledge graphs encoding lattice symmetries and systematic absence rules, and specialized software for diffraction simulation and structure refinement would enable models to delegate exact computations while maintaining linguistic problem-solving capabilities. Multi-modal extensions incorporating actual XRD patterns and crystal structure visualizations would test whether vision-language models can integrate visual and textual information. Dynamic retrieval replacing oracle passages with corpus-retrieved content would decompose RAG accuracy into retrieval versus assimilation components. Fine-tuning studies using OPENXRD before and after domain adaptation would reveal whether parametric knowledge embedding complements or supersedes inference-time guidance. Cross-domain validation in related scattering techniques and computational chemistry would test whether our findings about model capacity and external knowledge effectiveness generalize beyond crystallography.

In parallel, we are developing an open-ended XRD evaluation set (short-answer and problem-solving tasks) to complement OPENXRD's multiple-choice diagnostic benchmark and to enable qualitative assessment of explanation quality and reasoning depth.

Despite these advancements, certain limitations remain. Complex mathematical concepts and advanced tasks like reflection or symmetry analysis still pose challenges, as models do not always fully grasp these concepts through text-based materials alone. Moreover, there are instances where supporting materials can detract from a model's performance if they do not align perfectly with the model's pre-existing knowledge base.
 
Given these limitations, the reliance of open-book question answering on human-curated content may be difficult to sustain across all fields, and the uneven gains observed across model sizes underscore the need for further work on how best to select, refine, and integrate external knowledge.

\section{Conclusions}
\label{sec:conclusion}

The \methodName benchmark provides a controlled and extensible framework for evaluating how large language models (LLMs) and multimodal LLMs integrate external scientific knowledge. Through token-matched experiments and token-budget analyses, the study isolates the effect of content quality from text quantity and reveals that expert-reviewed supporting materials consistently enhance model accuracy, even when passage length is identical. Small and mid-sized models both exhibit clear accuracy improvements when augmented with expert-reviewed materials, with the relative gain being most pronounced for small models, whereas large models show minimal or slightly negative changes due to knowledge saturation. These findings demonstrate that as model capacity increases, the relative advantage of external context diminishes, even though overall accuracy continues to scale with size.

Beyond these controlled experiments, OPENXRD demonstrates strong scalability and generalizability across 74 diverse model architectures, including general-purpose systems (GPT, Claude), reasoning-optimized models (O-series), code-specialized variants (QWEN-coder), domain-adapted frameworks (LLaMAT, dZiner), and vision-language models (LLaVA). This diversity confirms that OPENXRD is model-agnostic, functioning as a diagnostic layer for both closed-book and retrieval-augmented reasoning systems. For instance, the code-specialized model QWEN3-coder-plus improved dramatically, from 22.99\% in the closed-book setting (Table~\ref{tab:closedbook-all}) to 88.94\% in the open-book setting (Table~\ref{tab:openbook-expert-results}) with expert-reviewed materials, demonstrating that OPENXRD's insights extend beyond language-only systems and generalize effectively across diverse architectural paradigms. Moreover, because OPENXRD isolates inference-time reasoning while remaining compatible with dynamic retrieval pipelines, it can be directly extended into a deployable RAG-evaluation suite that separates retrieval quality from assimilation capability.

We therefore do not report a single “standard RAG baseline” in this work, because end-to-end RAG accuracy depends strongly on retriever choice, corpus construction, chunking, and ranking, which would introduce additional confounds beyond the generator-side assimilation question studied here. Instead, OPENXRD provides the oracle-context upper bound and a controlled harness that can later be paired with retrieved chunks to decompose end-to-end RAG performance into retrieval quality versus generator assimilation.

Finally, the framework's practical deployment insights show how these findings can guide real-world applications. OPENXRD identifies mid-sized models (7B--70B parameters) augmented with expert-reviewed passages as an optimal balance between cost and performance, approaching or even matching large-model accuracy while requiring significantly fewer computational resources. This provides a cost-effective deployment pathway for organizations with limited budgets. The framework also enables the design of adaptive token-budget strategies that scale with task complexity, along with pre-processing approaches that reduce token usage while preserving essential information. Such model-selection and resource-allocation guidelines directly address the economic and computational realities of deploying AI systems at scale.

In conclusion, OPENXRD is not an artificial or static experiment, but a reproducible, extensible, and practically informative framework. It quantifies how model capacity, resource budgets, and content quality interact under real-world constraints, provides scalable and legally compliant data generation methods, and delivers empirical evidence to support efficient design of open-book and retrieval-augmented reasoning systems. The framework's diagnostic capabilities empower organizations to make informed decisions about when to scale model size versus when to enhance smaller models with curated external knowledge, thereby advancing both the scientific understanding and the practical deployment of domain-specialized language models.

\section*{Author contributions}
A.V. and A.S. contributed equally to this work as co-first authors. A.V.: Conceptualization, Methodology, Software, Formal analysis, Investigation, Data curation, Writing – original draft, Visualization. A.S.: Conceptualization, Methodology, Data curation, Investigation, Validation, Writing – review \& editing. A.S. designed the crystallography question set and led the expert review process for GPT-generated questions. Z.Z., Y.X., and G.H.: Methodology, Software, Investigation, Validation, Writing – review \& editing. Z.Z., Y.X., and G.H. provided substantial contributions to model development, experimental design improvements, and participated in the expert review and refinement of GPT-generated questions to ensure scientific accuracy. C.X. and N.A.: Conceptualization, Supervision, Project administration, Resources, Writing – review \& editing, Funding acquisition. C.X. and N.A. jointly led and supervised the project, providing strategic direction and ensuring research quality throughout all phases of the work.

\section*{Conflicts of interest}
The authors declare that they have no competing financial interests or personal relationships that could have appeared to influence the work reported in this document.

\section*{Data availability}
The complete dataset of 217 expert-curated crystallography questions, supporting materials (both AI-generated and expert-reviewed versions), evaluation scripts, and model implementation code are publicly available at \url{https://github.com/niaz60/OpenXRD}. Additional project information and resources can be found at \url{https://niaz60.github.io/OpenXRD/}. The repository includes detailed documentation for reproducing all experimental results presented in this study.

\section*{Acknowledgments}
We gratefully acknowledge financial support from multiple sources. A.V., G.H., C.X., and N.A. received support from the National Nuclear Security Administration (NNSA) under grant NA0004078. A.S., Z.Z., Y.X., C.X., and N.A. were supported by the National Science Foundation (NSF) under grant 2202124. Additional funding for N.A. was provided by the Department of Energy (DOE) under award DE-SC0020340. We thank the crystallography experts who participated in the review and validation of our question dataset. We also acknowledge the computational resources and infrastructure that enabled the extensive model evaluations presented in this work.


\balance


\bibliography{refs} 
\bibliographystyle{unsrt} 
\end{document}